\documentclass[runningheads]{llncs}

\usepackage[final,year=2026,ID=10155]{eccv}
\usepackage{eccvabbrv}

\usepackage{amsmath,amsfonts,bm}

\def\eqref#1{equation~\ref{#1}}

\def\1{\bm{1}}

\DeclareMathAlphabet{\mathsfit}{\encodingdefault}{\sfdefault}{m}{sl}
\SetMathAlphabet{\mathsfit}{bold}{\encodingdefault}{\sfdefault}{bx}{n}

\usepackage[breaklinks,colorlinks]{hyperref}
\usepackage{graphicx}
\usepackage{amsmath}
\usepackage{amsfonts}
\usepackage{amssymb}
\usepackage{subcaption}
\usepackage{booktabs}
\usepackage{multirow}
\usepackage{makecell}
\usepackage{colortbl}
\usepackage{xspace}
\usepackage{algorithm}
\usepackage{algorithmic}

\newcommand{\qwenTB}{\textsc{Qwen2.5-VL-3B}\xspace}
\newcommand{\qwenSB}{\textsc{Qwen2.5-VL-7B}\xspace}
\newcommand{\llavaSB}{\textsc{LLaVA-1.5-7B}\xspace}
\newcommand{\llavaffB}{\textsc{LLaVA-1.5-13B}\xspace}

\definecolor{ggreen}{rgb}{0.0, 0.0, 0.0}
\definecolor{rred}{rgb}{0.75, 0.0, 0.0}
\definecolor{bblue}{rgb}{0.13, 0.67, 0.8}

\newcommand{\goodmetric}[1]{{\color{ggreen} \textbf{#1}}}
\newcommand{\secondmetric}[1]{\underline{#1}}
\definecolor{ours}{rgb}{1.0, 1.0, 1.0}

\begin{document}

\title{Focusing by Contrastive Attention: Enhancing\texorpdfstring{\\}{ }VLMs' Visual Reasoning}
\titlerunning{Focusing by Contrastive Attention}

\author{Yuyao Ge$^{\clubsuit}$ \and Shenghua Liu$^{\clubsuit}$\thanks{Corresponding author.} \and Yiwei Wang$^{\triangle}$ \and Lingrui Mei$^{\clubsuit}$ \and \\ Baolong Bi$^{\clubsuit}$ \and Xuanshan Zhou$^{\clubsuit}$ \and Jiayu Yao$^{\clubsuit}$ \and \\ Jiafeng Guo$^{\clubsuit}$ \and Xueqi~Cheng$^{\clubsuit}$}
\authorrunning{Y. Ge et al.}

\institute{$^{\clubsuit}$Institute of Computing Technology, Chinese Academy of Sciences\\
$^{\triangle}$University of California, Merced\\
\email{\{geyuyao24z, liushenghua\}@ict.ac.cn}\\[9pt]
\raisebox{-2.5pt}{\includegraphics[height=1.1em]{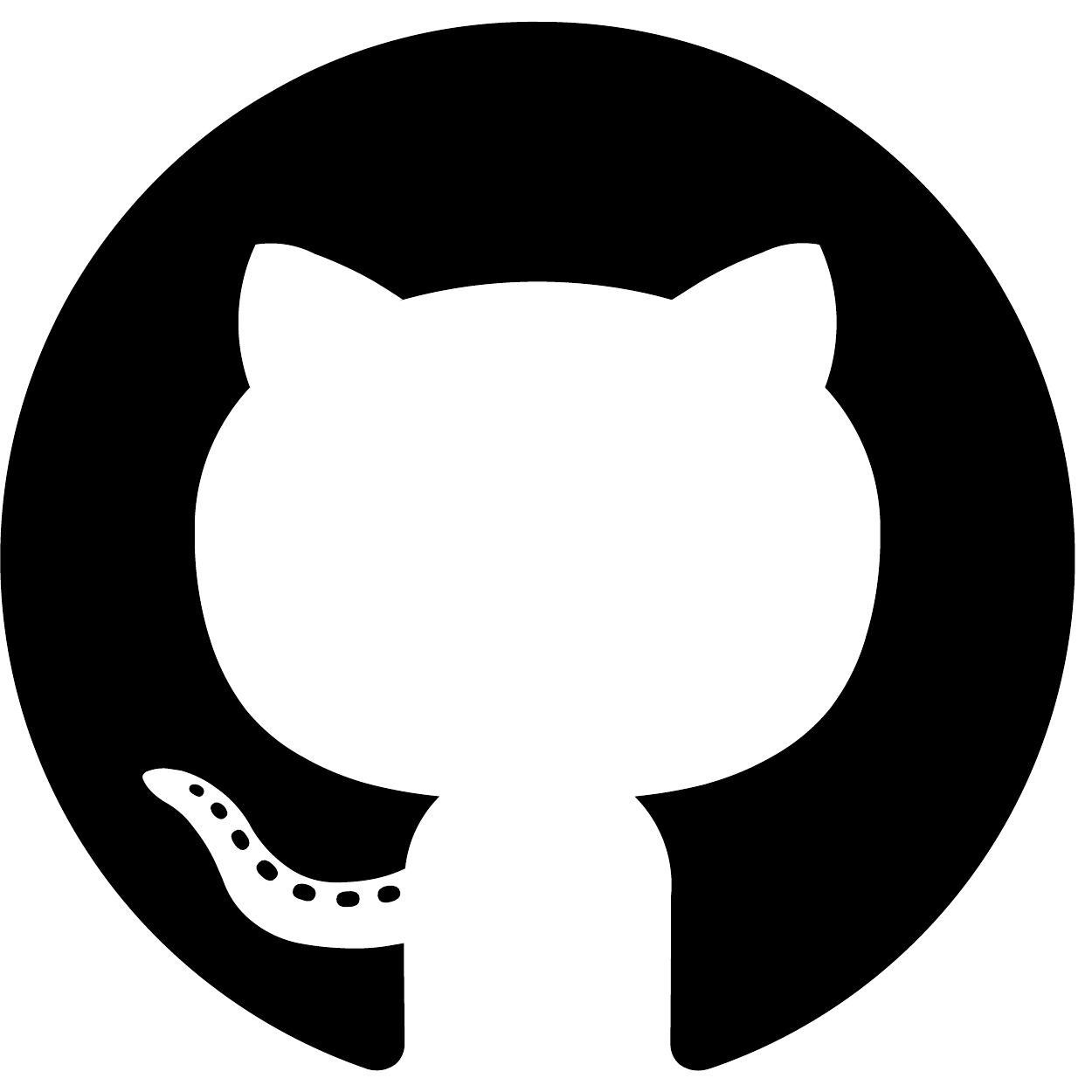}}~{\urlstyle{same}\url{https://github.com/YuyaoGe/CARVE}}}

\maketitle

\begin{abstract}
   Vision-Language Models (VLMs) have demonstrated remarkable success across diverse visual tasks, yet their performance degrades in complex visual environments.
   Existing enhancement approaches require additional training, rely on external segmentation tools, or operate at coarse-grained levels, overlooking VLMs' innate attention capabilities.
   To bridge this gap, we investigate VLMs' attention patterns and discover that: (1) visual complexity strongly correlates with attention entropy, negatively impacting reasoning performance;
   (2) attention progressively refines from global scanning in shallow layers to focused convergence in deeper layers, with the degree of convergence determined by visual complexity;
   (3) theoretically, under a multiplicative decomposition assumption, we show that contrasting attention maps between general and task-specific queries approximately separates visual signal into semantic and visual noise components.
   Building on these insights, we propose \textbf{C}ontrastive \textbf{A}ttention \textbf{R}efinement for \textbf{V}isual \textbf{E}nhancement (\textbf{CARVE}), a training-free method that extracts task-relevant visual signals through attention contrasting at the pixel level.
   Experiments on seven benchmarks show that CARVE improves performance on visual perception tasks, with notable gains on both recent and earlier-generation models.
   Our analysis reveals how visual complexity affects attention mechanisms and demonstrates an effective strategy for improving visual reasoning through attention contrasting.

\end{abstract}

\section{Introduction}
\label{sec:introduction}

Vision-Language Models (VLMs) have shown impressive capabilities across diverse tasks~\cite{kimiteam2026kimi, radford2021learning, jia2021scaling, alayrac2022flamingo}. However, in human vision, complex visual features frequently divert attention from task-relevant regions~\cite{treisman1980featureintegration}. Given this cognitive parallel, a question arises: \textit{Similarly, do complex images interfere with VLMs' attention mechanisms, making it difficult for them to focus on task-relevant regions?}

To answer this question, we investigate the relationship between visual complexity and attention patterns via quantitative experiments.
Specifically, we define visual complexity along texture and color dimensions, revealing a significant positive correlation between both factors and attention entropy.
Our analysis also shows that attention entropy negatively correlates with accuracy on visual reasoning tasks.
Through this two-stage analysis, we find that \textbf{complex visual information impairs VLMs' reasoning performance via attention dispersion} (Sect.~\ref{sec:visual_complexity_attention}).

\begin{figure}[t] 
    \centering
    \includegraphics[width=\textwidth]{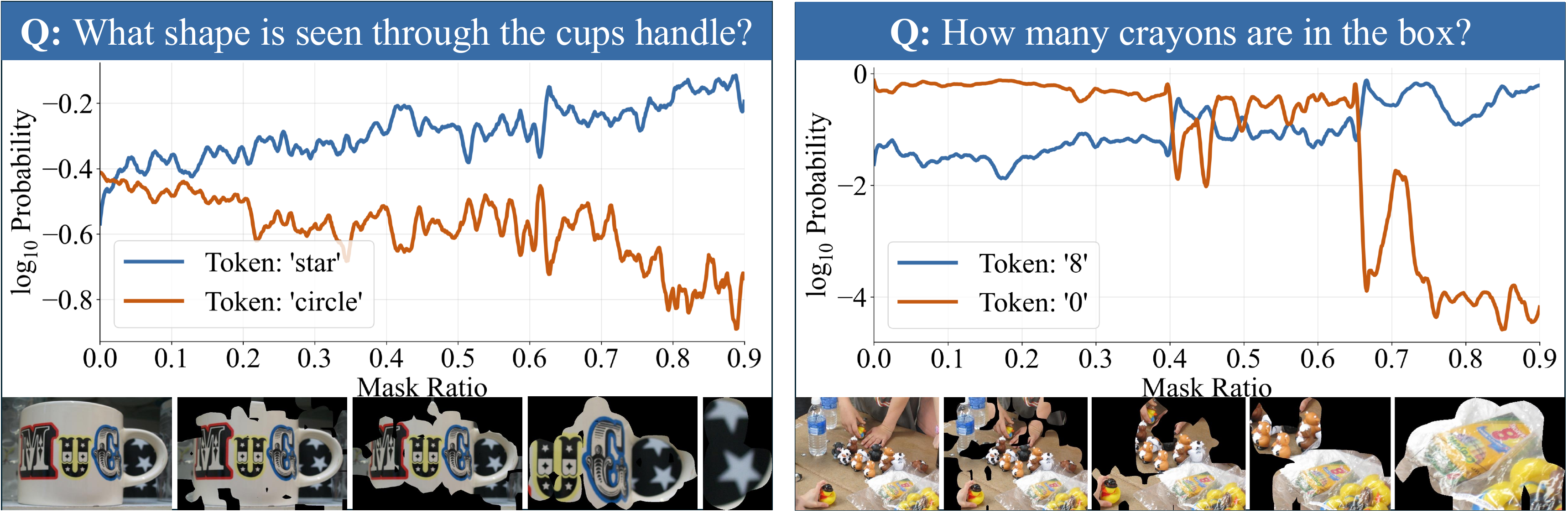}
    \caption{
    The effect of manually progressive masking on candidate tokens probabilities predicted by \qwenTB.
    The x-axis represents mask ratio and y-axis shows $\log_{10}$ probability.
    }
    \label{fig:effectiveness_demo}
\end{figure}

Based on these findings, and inspired by ViCrop~\cite{zhang2025mllms}, we conduct a preliminary experiment on TextVQA~\cite{singh2019vqa} by progressively masking background regions, then cropping and magnifying task-relevant areas.
Fig.~\ref{fig:effectiveness_demo} presents two representative samples where cluttered environments initially cause incorrect predictions.
As more background is masked, correct token probability surpasses incorrect probability, validating that \textbf{masking visual noise can improve correct token probability}.

To automate the visual noise masking process, we exploit contrasting attention maps between general instructions and task-specific questions to distinguish semantic signal from visual noise.
To this end, we propose \textbf{C}ontrastive \textbf{A}ttention \textbf{R}efinement for \textbf{V}isual \textbf{E}nhancement (\textbf{CARVE}), a training-free contrastive method for visual extraction.
By masking with contrastive attention maps, CARVE crops and magnifies semantic regions to focus on essential semantic signal (Sect.~\ref{sec:method}).

\section{Related Work}
\label{sec:related_work}

\paragraph{Contrastive Learning in LLMs.}
Liu~\etal~\cite{liu2024learning} pioneered contrastive objectives for distilling supervision from LLMs.
To address hallucination, Jiang~\etal~\cite{jiang2024hallucination} employ hallucinated text as hard negatives while Zhang~\etal~\cite{zhang2024truthx} apply contrastive learning in hidden representations to suppress hallucinations.
Zhai~\etal~\cite{zhai2025parameteraware} trace critical transmission paths across all layers, treating less important pathways as negatives, which Pan~\etal~\cite{pan2024unified} further adapted to multimodal LLMs through UniKE's semantic-truthfulness space disentanglement.
Departing from embedding-space modifications, DeCK~\cite{bi2025decoding} shifts contrastive logic to the decoding stage by comparing token probabilities with and without injected knowledge.

\paragraph{Attention-based LLM Optimization.}
Training-free approaches prune redundant visual tokens~\cite{chen2024image,ma2024eemllm,acharya2025star,chen2025pis} or broadcast vision-focused attention heads to sibling heads~\cite{zhang2024seeing} to accelerate inference and reduce hallucinations.
Training-based methods ground and zoom into task-relevant regions: CogCoM~\cite{qi2024cogcom} trains a chain-of-manipulation pipeline for step-by-step visual reasoning with grounding and zooming, CogAgent~\cite{hong2024cogagent} employs dual encoders for high-resolution understanding, Ferret-UI~\cite{you2024ferretui} splits screens into sub-images for fine-grained grounding, VistaWise~\cite{fu2025vistawise} finetunes an object detector to isolate task-relevant entities from visual clutter for embodied agents, and Gkartzonika~\etal~\cite{gkartzonika2022learning} train attention mechanisms on frozen classifiers to produce visual explanation maps.
Complementary analyses reveal that models know where to look even when answering incorrectly~\cite{zhang2025mllms} and that position bias affects multimodal retrieval~\cite{yao2025spotlight}.
Unlike these methods, our approach contrasts attention maps to distinguish semantic pixels from noise at the pixel level, requiring no training.

\section{Failure to Focus: Phenomenon, Mechanism, and Consequence}
\label{sec:visual_complexity_attention}

\subsection{Phenomenon: Under What Conditions Visual Focus Fails}
\label{subsec:identifying_problem}

\begin{figure}[t] 
    \centering
    \includegraphics[width=\textwidth]{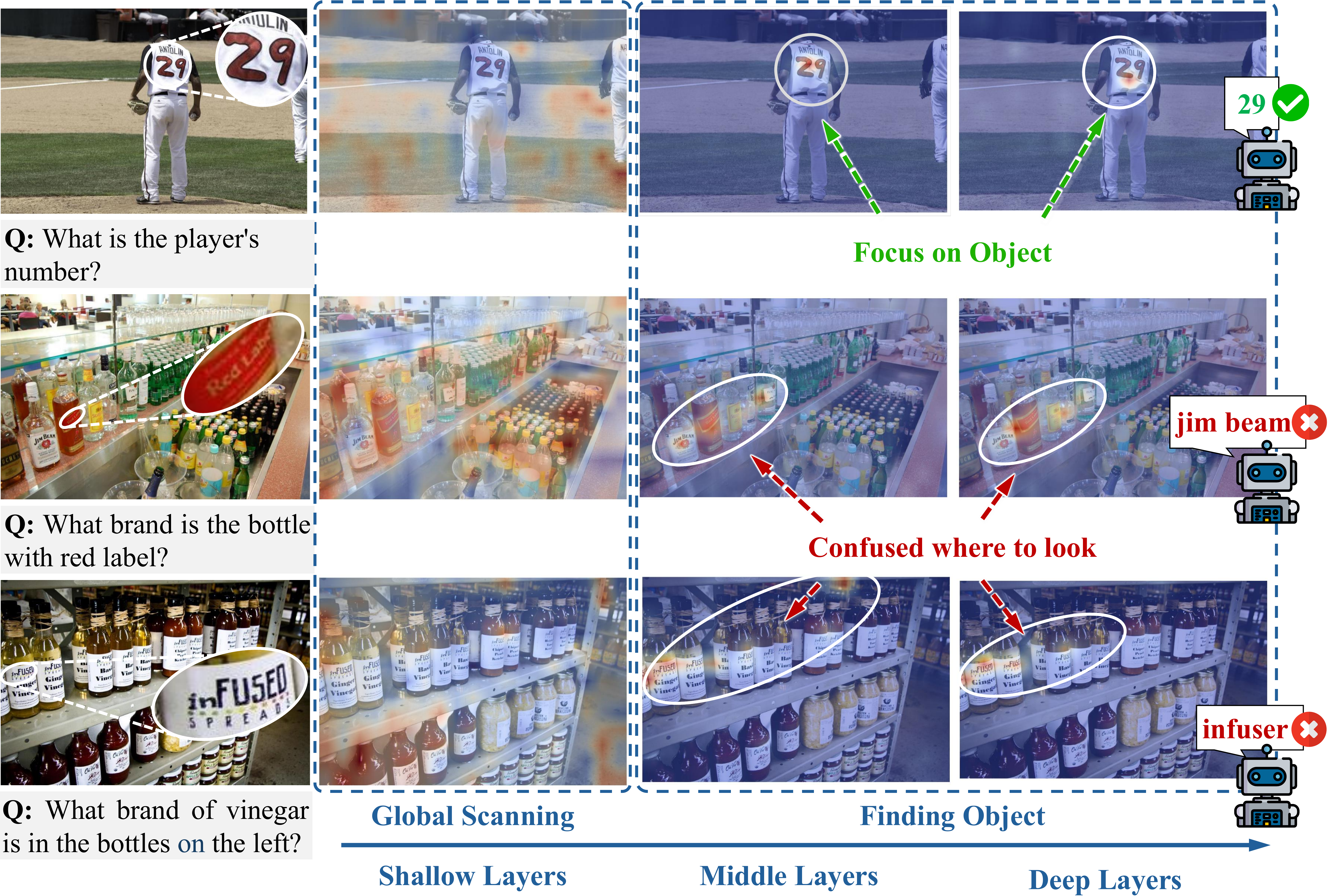}
    \caption{Attention maps across different layers during inference. Each row represents a visual question-answering task: row 1 shows a simple scene with clear targets, while rows 2-3 depict complex scenes with dense textures and multiple similar objects. From left to right, the columns display: input images, shallow layer attention, middle layer attention, and deep layer attention.}
    \label{fig:problem_discovery}
\end{figure}

We visualize attention maps of \textsc{Qwen2.5-VL-3B-Instruct} on TextVQA (Fig.~\ref{fig:problem_discovery}).
Attention progressively refines from global scanning in shallow layers to focused convergence in deep layers, with convergence varying across images.

Visual complexity also influences attention convergence. In simple scenes with clear targets and minimal distractors, attention successfully narrows to task-relevant regions as layers deepen (consistent with ViCrop~\cite{zhang2025mllms}).
Conversely, in complex scenes with rich textures and colors, attention remains more dispersed even in deep layers.
As indicated by the annotation ``Confused where to look'' in Fig.~\ref{fig:problem_discovery}, this attention dispersion leads to reasoning failures.

These observations motivate two quantitative investigations: Sect.~\ref{subsec:quantifying_complexity} examines whether visual complexity disperses attention, and Sect.~\ref{subsec:entropy_performance_connection} links attention dispersion to performance degradation.

\subsection{Mechanism: Visual Complexity and Attention Entropy}
\label{subsec:quantifying_complexity}
In Fig.~\ref{fig:problem_discovery}, images in rows 2-3 differ from row 1 by displaying numerous colorful bottles, containing significantly more textures and colors.
Therefore, we decompose visual complexity into two dimensions: \textbf{texture} and \textbf{color}, and investigate their respective impacts on attention.

We define the texture complexity and the color complexity as follows:

\textbf{Texture Complexity.} Let $\mathcal{I} \in \mathbb{R}^{H \times W \times 3}$ denote an input image. We define the texture complexity $\mathcal{T}_c(\mathcal{I})$ using Canny edge detection~\cite{canny1986computational}, where $\mathcal{E}(\mathcal{I}) \in \{0,1\}^{H \times W}$ represents the resulting binary edge map. The texture complexity is then defined as:

\begin{equation}
\mathcal{T}_c(\mathcal{I}) = \frac{1}{HW} \sum_{i=1}^{H} \sum_{j=1}^{W} \mathcal{E}(\mathcal{I})_{ij} = \frac{\|\mathcal{E}(\mathcal{I})\|_1}{HW} \in [0, 1]
\end{equation}

\textbf{Color Complexity.}
Let $\zeta_{ij} = \text{Hue}(\Psi_{RGB \rightarrow HSV}(\mathcal{I}_{ij}))$ denote the hue value at pixel $(i,j)$ after applying the RGB to HSV transformation operator $\Psi$. The color complexity is then defined as:
\begin{equation}
\mathcal{C}_c(\mathcal{I}) = -\frac{1}{\ln B} \sum_{b=0}^{B-1} \rho_b \ln \rho_b, \quad \text{where } \rho_b = \frac{n_b}{HW}, \quad n_b = |\{(i,j) : \zeta_{ij} = b\}|
\end{equation}
with $B = 180$ hue bins, yielding $\mathcal{C}_c \in [0, 1]$ where higher values indicate greater color diversity.

\begin{figure}[t]
    \centering
    \newcommand{\figheight}{1.9cm}

    \begin{subfigure}[b]{0.23\textwidth}
        \includegraphics[width=\textwidth, height=\figheight, keepaspectratio=false]{figure/16_original_image.pdf}
        \label{fig:16_original}
    \end{subfigure}
    \hfill
    \begin{subfigure}[b]{0.23\textwidth}
        \includegraphics[width=\textwidth, height=\figheight, keepaspectratio=false]{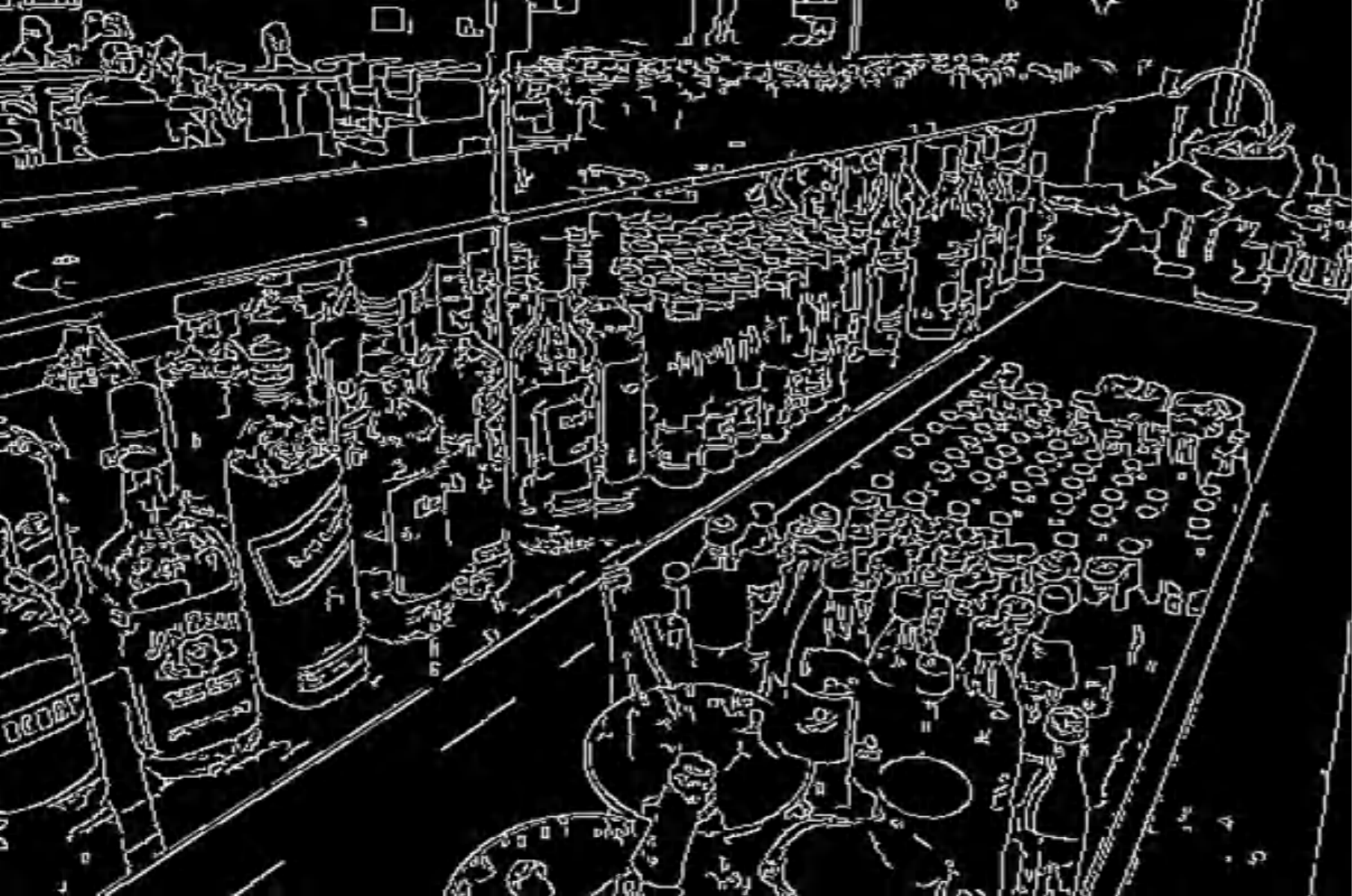}
        \label{fig:16_edge}
    \end{subfigure}
    \hfill
    \begin{subfigure}[b]{0.23\textwidth}
        \includegraphics[width=\textwidth, height=\figheight, keepaspectratio=false]{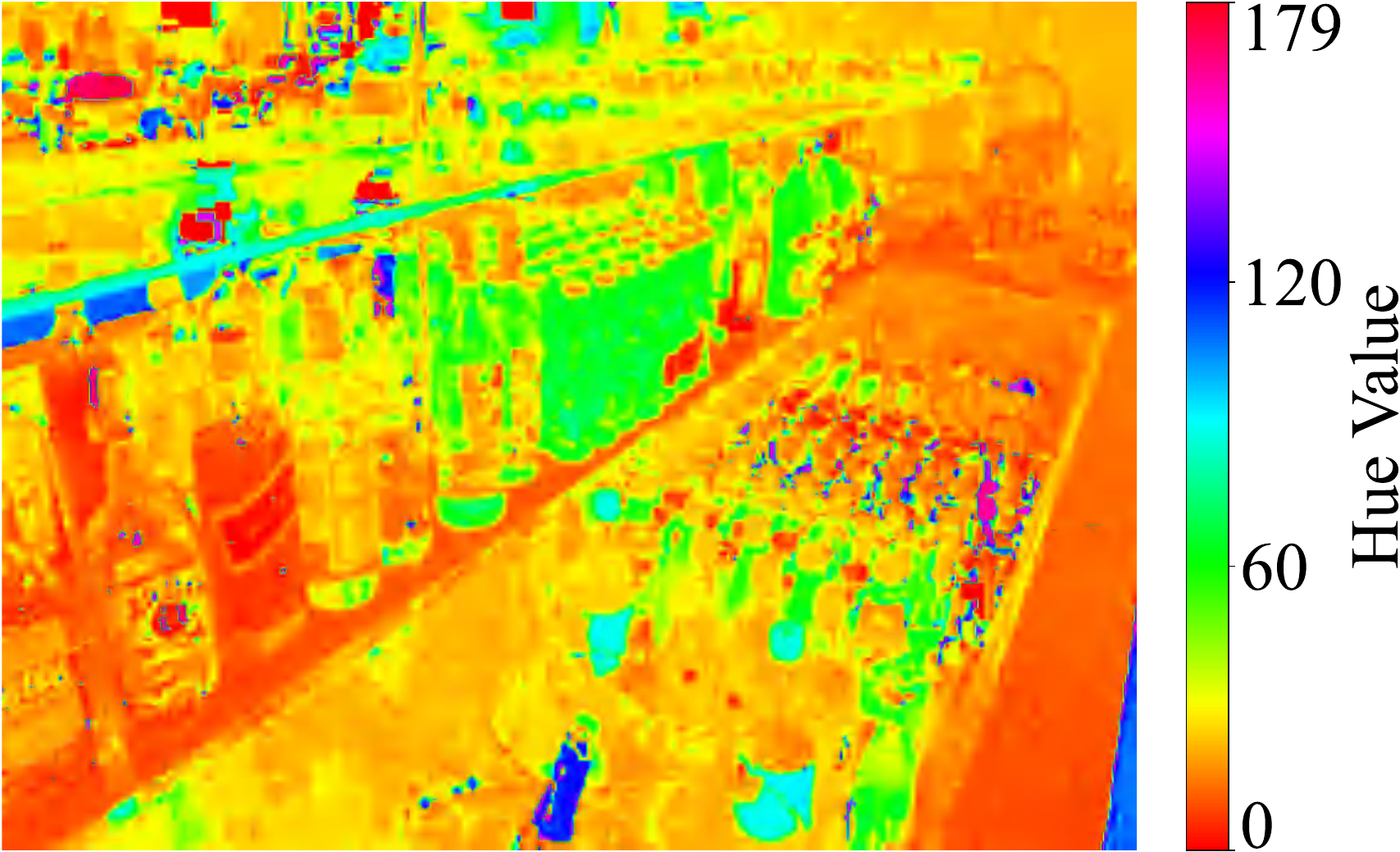}
        \label{fig:16_hue_dist}
    \end{subfigure}
    \hfill
    \begin{subfigure}[b]{0.23\textwidth}
        \includegraphics[width=\textwidth, height=\figheight, keepaspectratio=false]{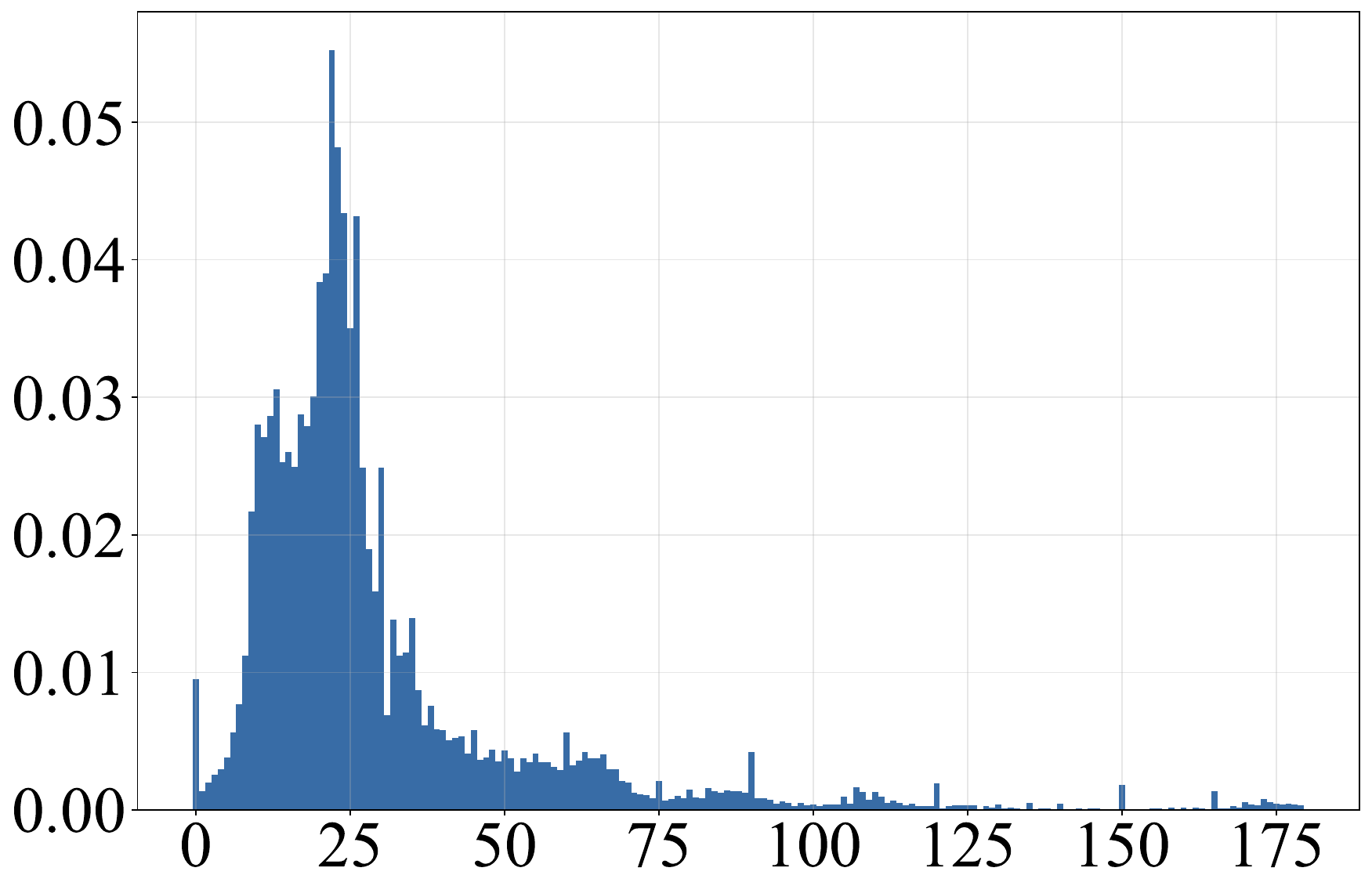}
        \label{fig:16_hue_hist}
    \end{subfigure}

    \begin{subfigure}[b]{0.23\textwidth}
        \includegraphics[width=\textwidth, height=\figheight, keepaspectratio=false]{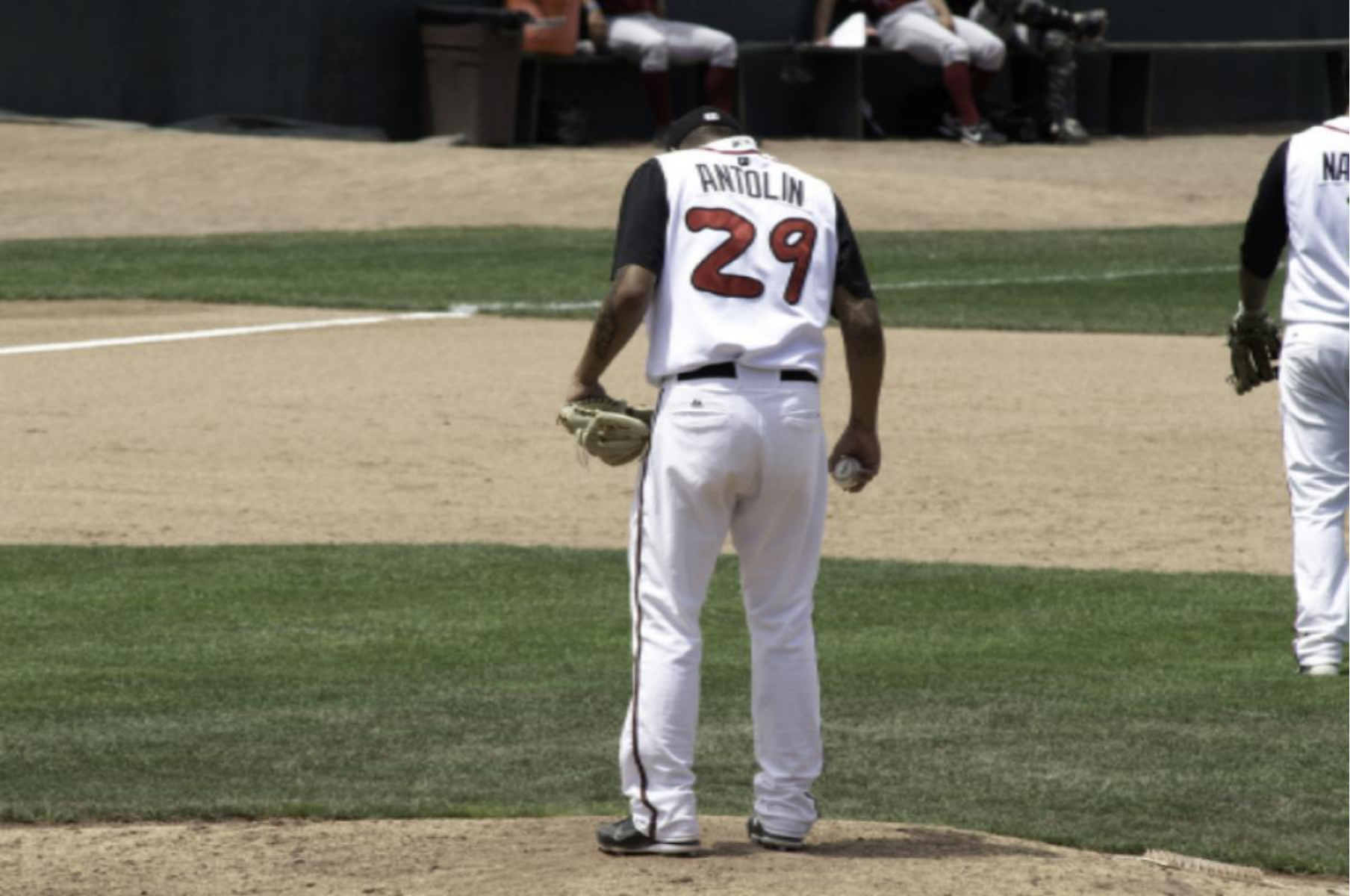}
        \caption{Original}
        \label{fig:2550_original}
    \end{subfigure}
    \hfill
    \begin{subfigure}[b]{0.23\textwidth}
        \includegraphics[width=\textwidth, height=\figheight, keepaspectratio=false]{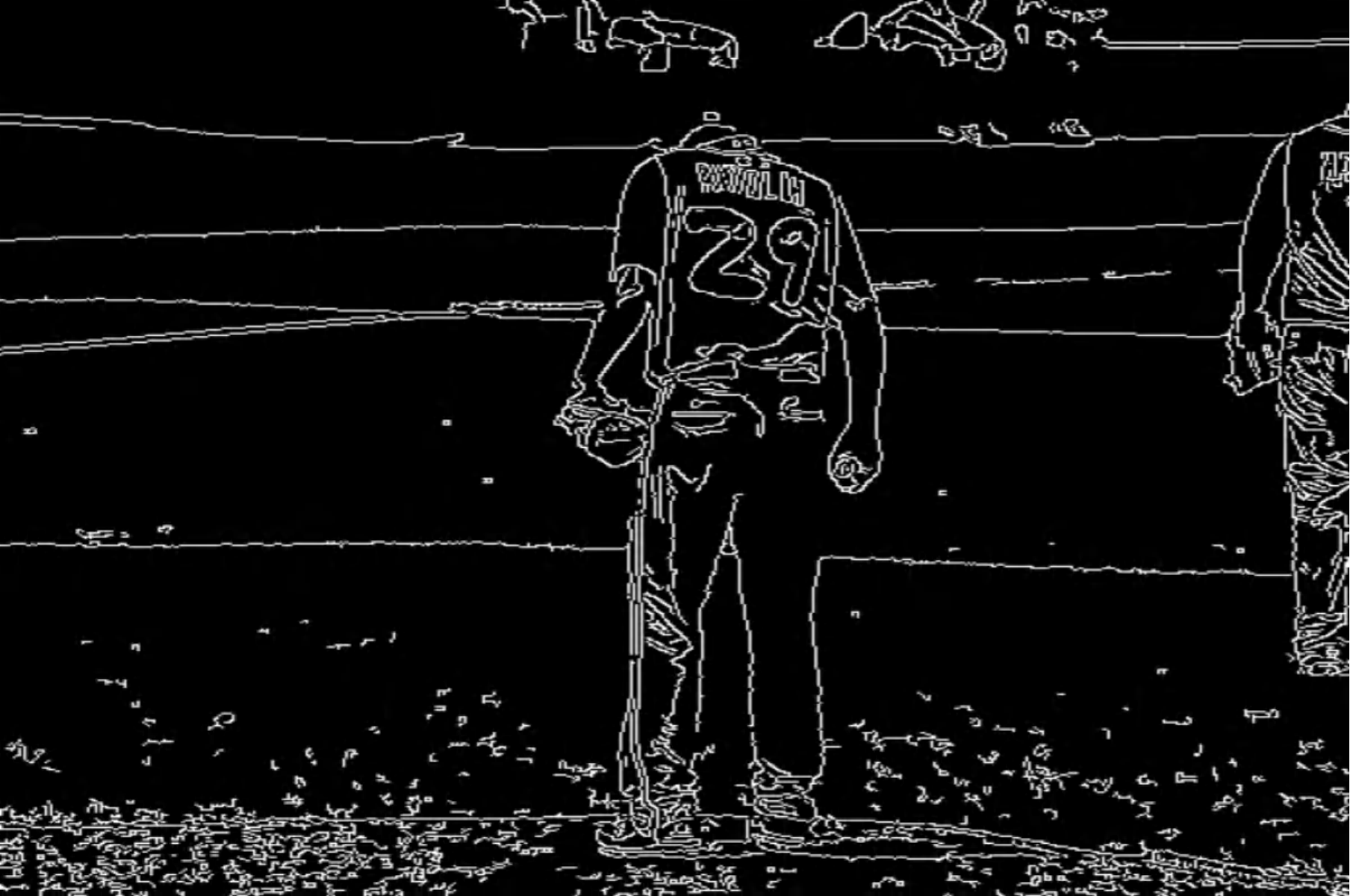}
        \caption{Edge detection}
        \label{fig:2550_edge}
    \end{subfigure}
    \hfill
    \begin{subfigure}[b]{0.23\textwidth}
        \includegraphics[width=\textwidth, height=\figheight, keepaspectratio=false]{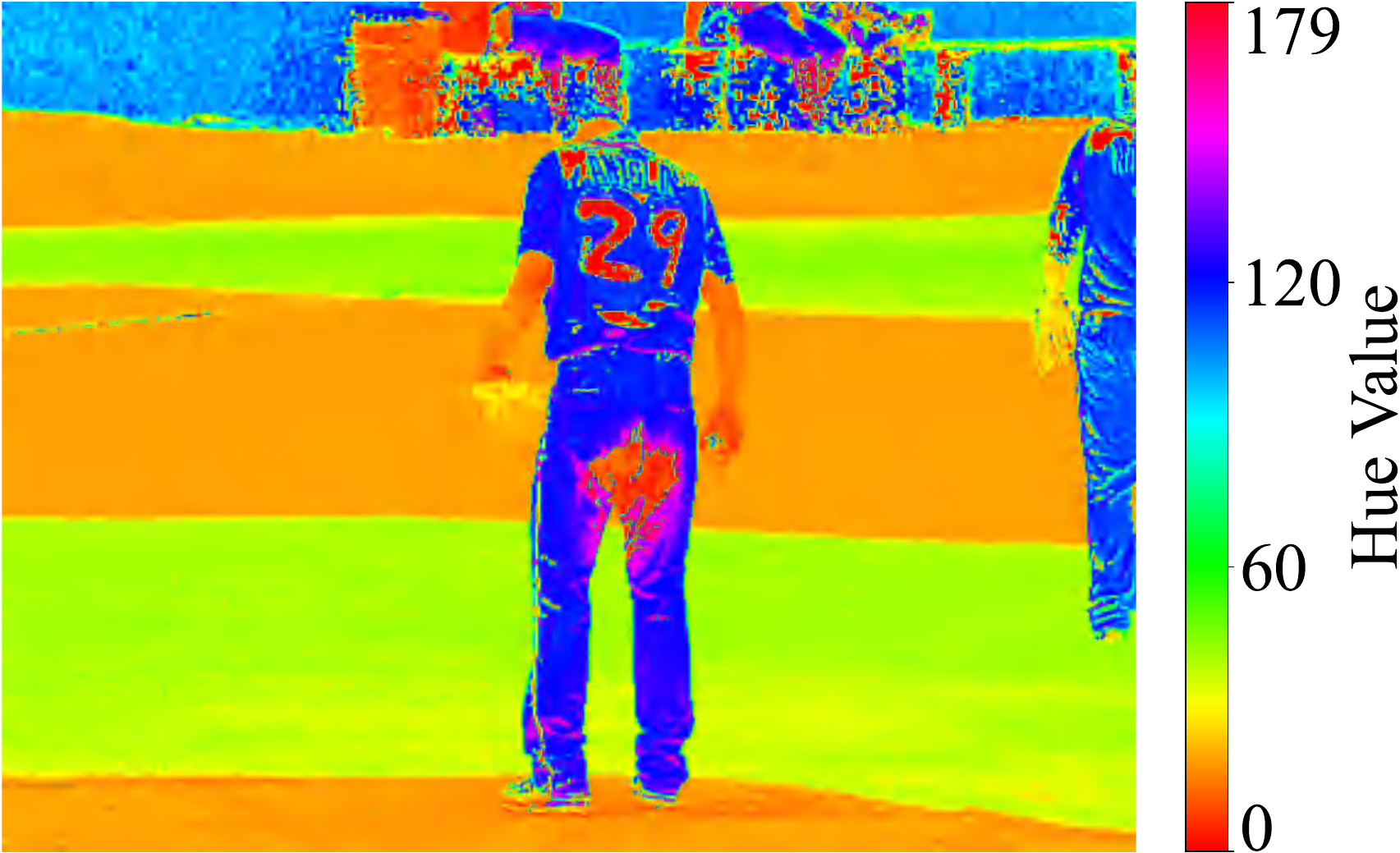}
        \caption{Hue distribution}
        \label{fig:2550_hue_dist}
    \end{subfigure}
    \hfill
    \begin{subfigure}[b]{0.23\textwidth}
        \includegraphics[width=\textwidth, height=\figheight, keepaspectratio=false]{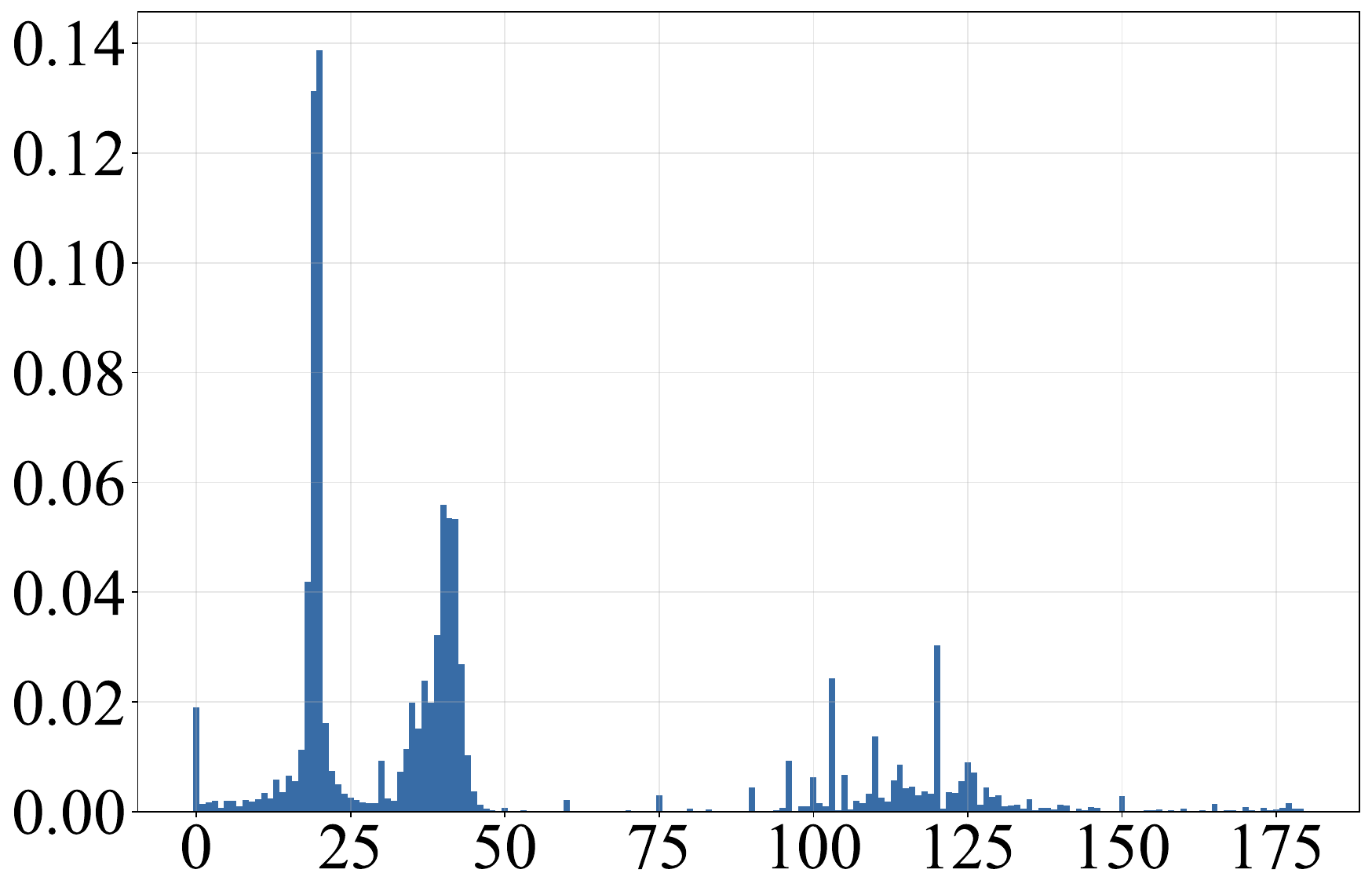}
        \caption{Hue histogram}
        \label{fig:2550_hue_hist}
    \end{subfigure}

    \caption{Visualization of texture and color complexity analysis. Each row represents a sample image: (a) Original image $\mathcal{I}$, (b) Canny edge map $\mathcal{E}(\mathcal{I})$ for texture complexity $\mathcal{T}_c$, (c) Spatial hue distribution $\zeta$ in HSV space, and (d) Hue statistic (x-axis: hue value, y-axis: ratio).}

    \label{fig:complexity_analysis_samples}
\end{figure}

\begin{figure}[t]
    \centering
    \begin{subfigure}[b]{0.48\textwidth}
        \includegraphics[width=\textwidth]{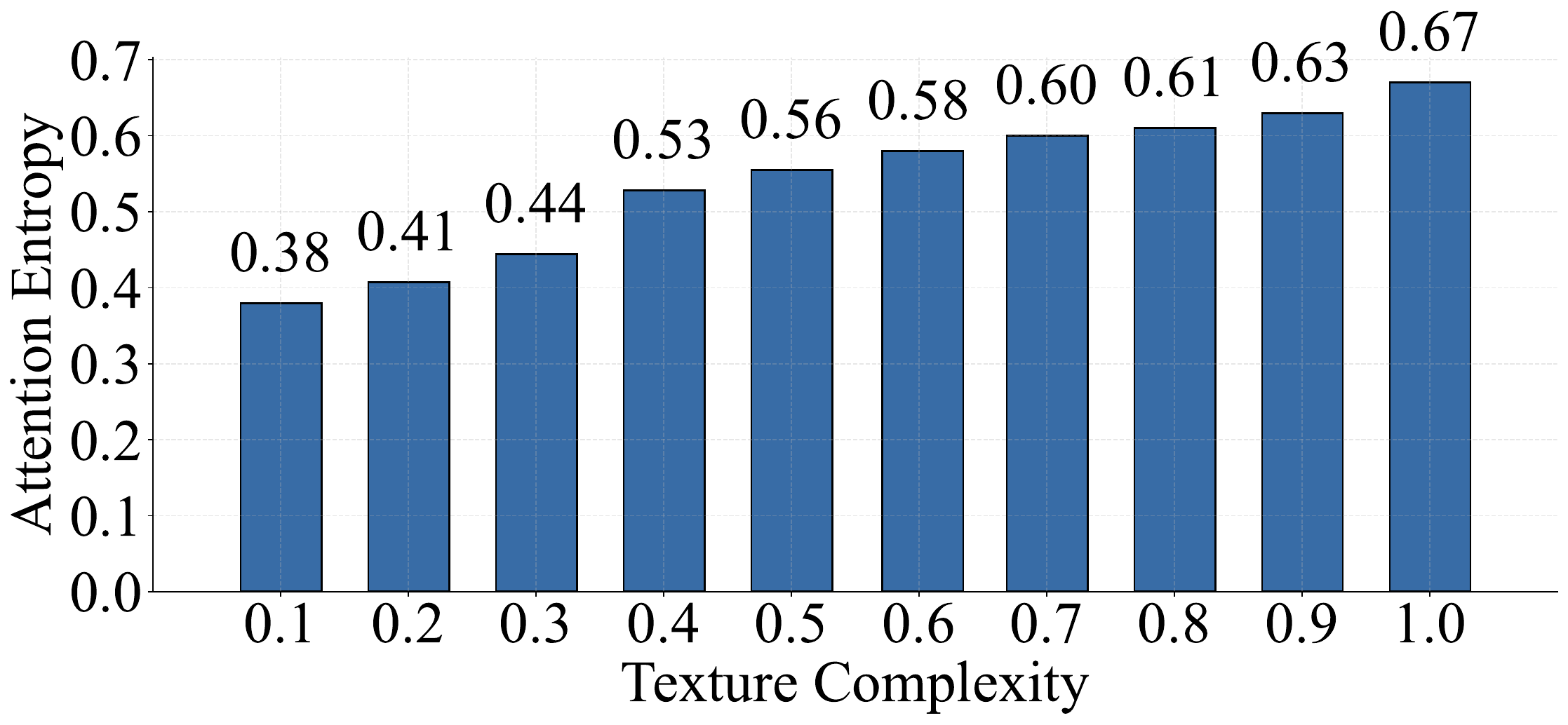}
        \caption{Texture-entropy correlation}
        \label{fig:texture_correlation}
    \end{subfigure}
    \hfill
    \begin{subfigure}[b]{0.48\textwidth}
        \includegraphics[width=\textwidth]{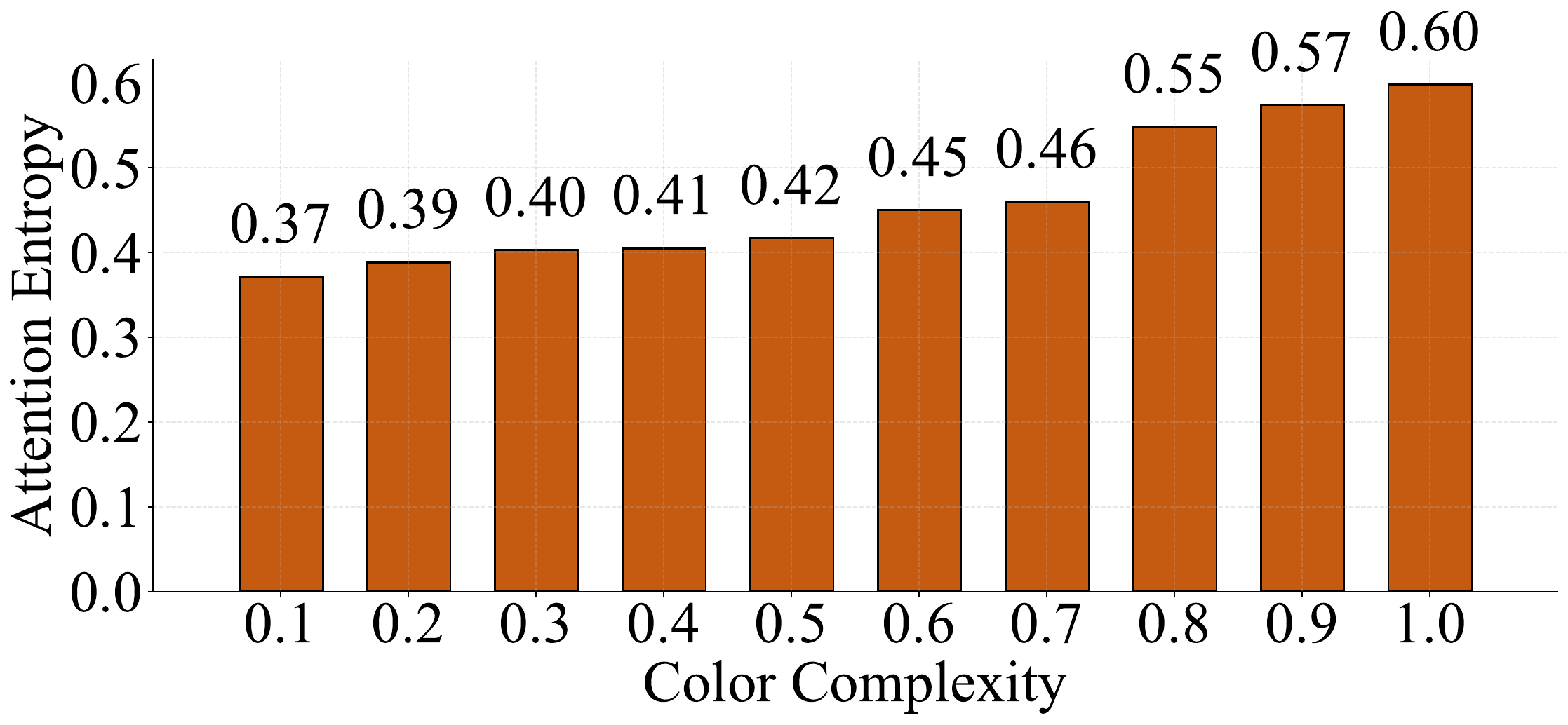}
        \caption{Color-entropy correlation}
        \label{fig:color_correlation}
    \end{subfigure}
    \caption{Correlation analysis between visual complexity and attention entropy. Both attention entropy and complexity are normalized to the $[0, 1]$, divided into intervals of 0.1, and the average attention entropy is calculated within each interval.
    }
    \label{fig:complexity_correlation}
\end{figure}

Fig.~\ref{fig:complexity_analysis_samples} validates our complexity metrics. The first row shows high texture complexity with dense edge networks in $\mathcal{E}(\mathcal{I})$ and diverse color distribution across the hue spectrum. In contrast, the second row exhibits minimal edge density and concentrated hue values, indicating lower complexity scores. We next quantitatively investigate the correlation between complexity metrics and attention distribution.

Inspired by Yao~\etal~\cite{yao2025spotlight}, we employ Shannon entropy~\cite{shannon1948mathematical} to quantify attention distribution across visual tokens.
Let $N_v$ denote the number of visual tokens in the model's representation. We denote the attention map as $A_{l,t}^{(Q)} \in \mathbb{R}^{N_v}$, where $l$ indicates the layer index, $t$ the generation time step, and $Q$ the input question.
For entropy analysis, we focus on the final generation step $t_{\text{end}}$ and define the overall attention entropy $\overline{\mathcal{H}}$ as:

\begin{equation}
\overline{\mathcal{H}} = \frac{1}{|\mathcal{L}|} \sum_{l \in \mathcal{L}} \mathcal{H}(A_{l,t_{\text{end}}}^{(Q)}) = \frac{1}{|\mathcal{L}|} \sum_{l \in \mathcal{L}} \left(-\sum_{i=1}^{N_v} a_{l,t_{\text{end}},i} \ln a_{l,t_{\text{end}},i}\right)
\end{equation}

where $\mathcal{L} = [L_{\text{start}}, L_{\text{end}}]$ represents the layer range under consideration,
and $a_{l,t,i}$ denotes the attention weight for the $i$-th visual token. Higher entropy indicates more dispersed attention, while lower entropy indicates more concentrated focus.

Fig.~\ref{fig:complexity_correlation} presents the correlation between visual complexity and attention entropy. Both texture complexity (Fig.~\ref{fig:texture_correlation}) and color complexity (Fig.~\ref{fig:color_correlation}) exhibit strong positive linear relationships with attention entropy.
This monotonic trend indicates that \textbf{complex visual features lead to dispersed attention patterns in VLMs}.

\begin{figure}[t]
    \centering
    \begin{subfigure}[b]{0.48\textwidth}
        \includegraphics[width=\textwidth]{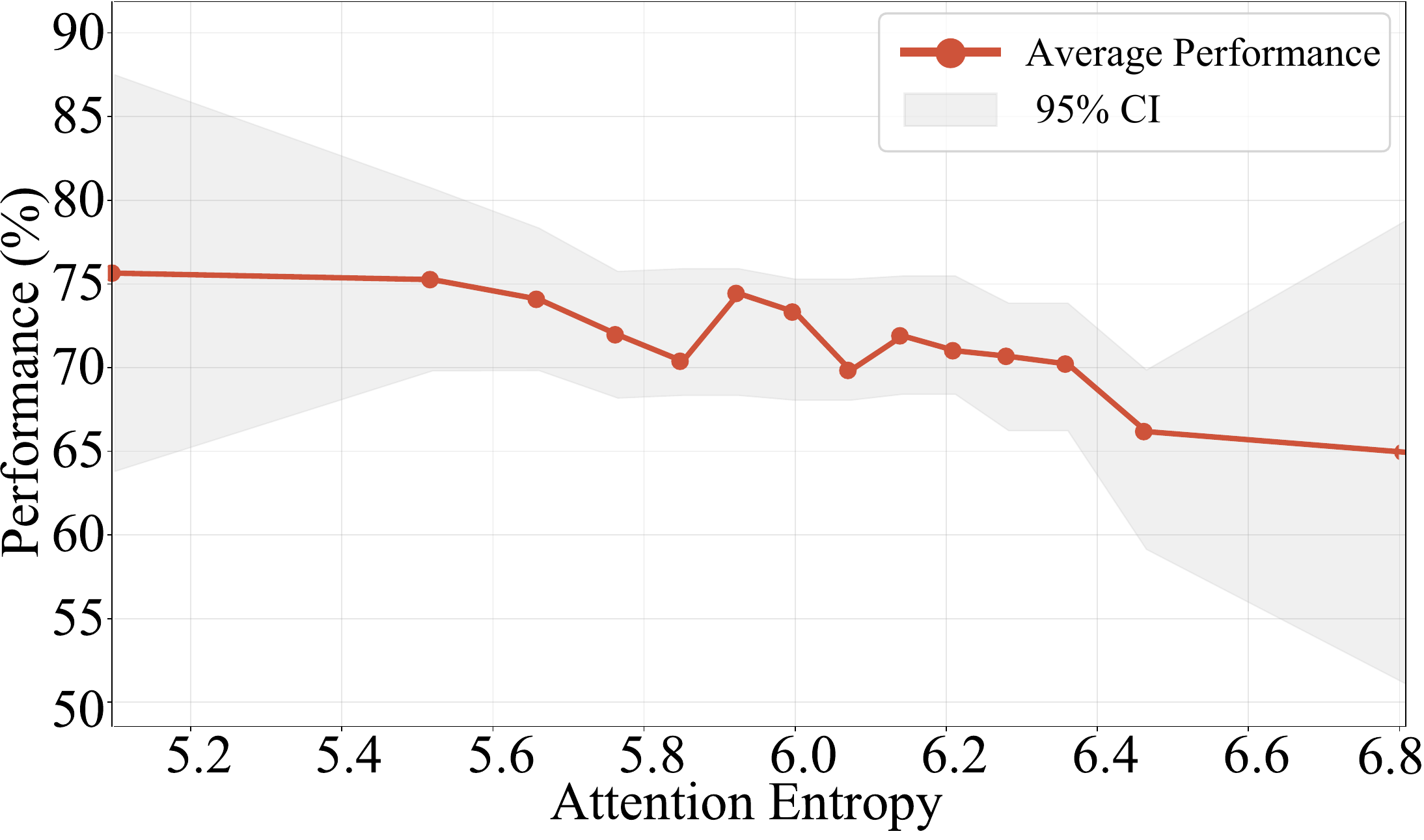}
        \caption{Entropy-accuracy correlation}
        \label{fig:entropy_performance_sub}
    \end{subfigure}
    \hfill
    \begin{subfigure}[b]{0.48\textwidth}
        \includegraphics[width=\textwidth]{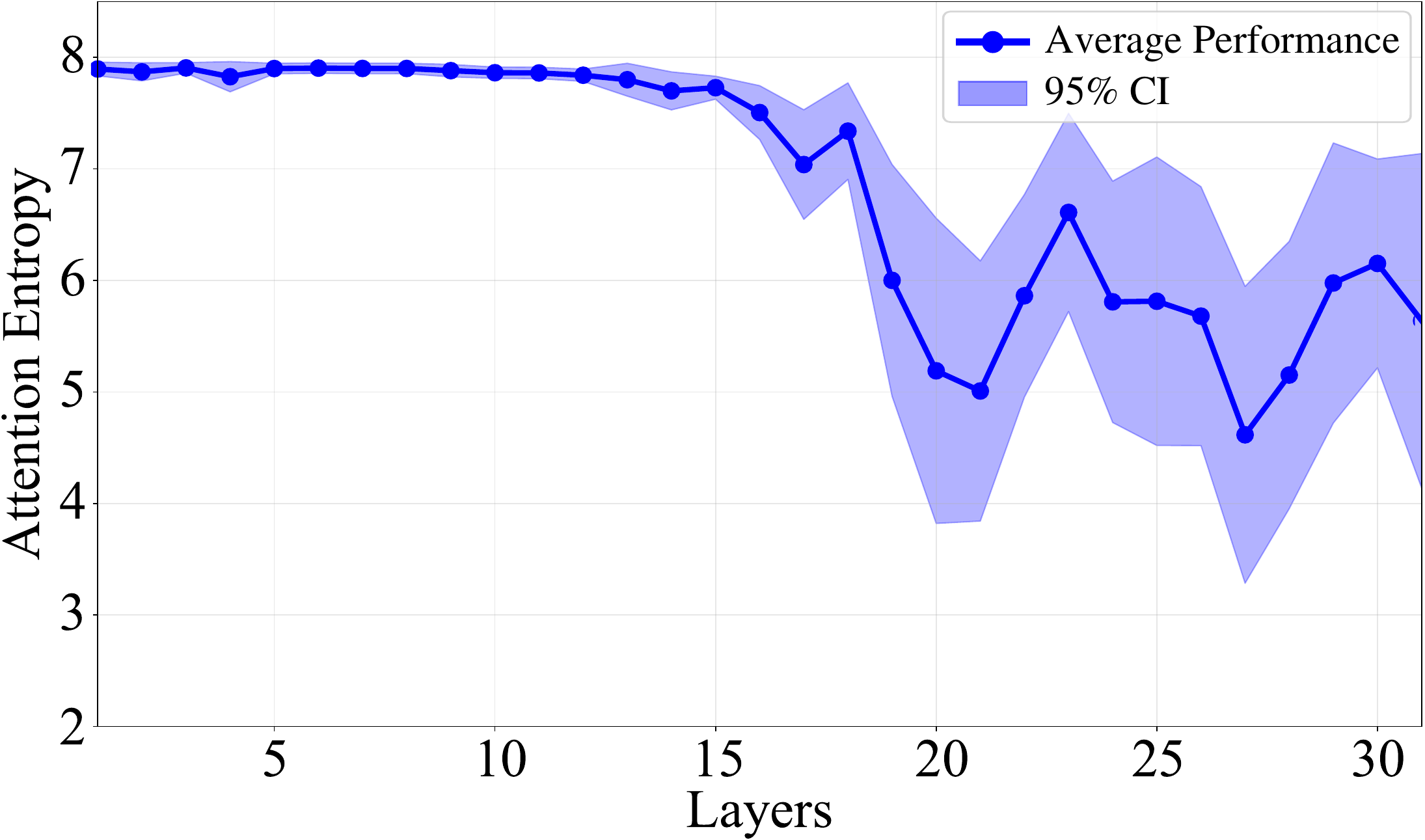}
        \caption{Layer-wise entropy correlation}
        \label{fig:attention_evolution_sub}
    \end{subfigure}
    \caption{Attention entropy's correlation with accuracy and its evolution across layers. Shaded regions indicate 95\% confidence intervals computed as $\bar{x} \pm t_{0.975,n-1} \cdot s/\sqrt{n}$.
    (a) Shows accuracy for samples grouped by overall attention entropy $\overline{\mathcal{H}}$.
    (b) Displays mean entropy across $N$ samples per layer as $\frac{1}{N}\sum_{i=1}^{N}\mathcal{H}(A_{l,t_{\text{end}}}^{(Q,i)})$, where $A_{l,t_{\text{end}}}^{(Q,i)}$ is sample $i$'s attention at the final generation step.
    }
    \label{fig:entropy_performance}
\end{figure}

\subsection{Consequence: How Dispersed Attention Impairs Performance}
\label{subsec:entropy_performance_connection}
Fig.~\ref{fig:entropy_performance}(a) reveals a strong negative correlation: as attention entropy increases from 5.1 to 6.8, accuracy drops from approximately 76\% to 65\%, suggesting that higher attention dispersion is associated with lower visual reasoning accuracy.

To investigate the hierarchical evolution of attention entropy, we present mean entropy and its distribution across layers in Fig.~\ref{fig:entropy_performance}(b).
The results show two characteristics: (1) attention entropy monotonically decreases with layer depth, consistent with Fig.~\ref{fig:problem_discovery}.
(2) The 95\% confidence intervals progressively widen with increasing depth, indicating enhanced inter-sample variability.
For samples with clear visual targets, deep layers achieve highly concentrated attention.
In contrast, for noisy samples, the model maintains dispersed attention patterns even in deep layers.

\section{\textbf{C}ontrastive \textbf{A}ttention \textbf{R}efinement for \textbf{V}isual \textbf{E}nhancement}
\label{sec:method}

\subsection{Theoretical Foundation: Attention Decomposition}
\label{subsec:theoretical_foundation}
Given that visual complexity causes attention dispersion and performance degradation (Sect.~\ref{sec:visual_complexity_attention}), we introduce an attention decomposition to extract pure semantic signal.

\textbf{Assumption 1 (Attention Decomposition):} Attention distributions are influenced by inherent visual noise of the image and task-related semantic signal. We model the attention map $A_{l,t}^{(Q)}(\mathcal{I})$ as:

\begin{equation}
A_{l,t}^{(Q)}(\mathcal{I}) = \mathcal{F}_{\text{vis}}(\mathcal{I}) \otimes \mathcal{F}_{\text{sem}}(Q, \mathcal{I})
\end{equation}

where $\mathcal{F}_{\text{vis}}(\mathcal{I}) \in \mathbb{R}^{N_v}$ captures image-inherent visual noise, $\mathcal{F}_{\text{sem}}(Q, \mathcal{I}) \in \mathbb{R}^{N_v}$ captures task-related semantic signal, and $\otimes$ denotes the Hadamard product.
We adopt multiplicative rather than additive decomposition because visual noise acts as a spatially varying gain on attention weights; element-wise division normalizes heterogeneous noise magnitudes uniformly, whereas subtraction leaves scale-dependent residuals.
Empirically, replacing the multiplicative contrast with an additive alternative $\hat{A}_i {=} A_i^{(Q)} {-} A_i^{(G)}$ reduces the relative gain on TextVQA from $+$9.33\% to $+$4.53\%, supporting the benefit of scale-invariant normalization.

\begin{figure}[t]
    \centering
    \includegraphics[width=\textwidth]{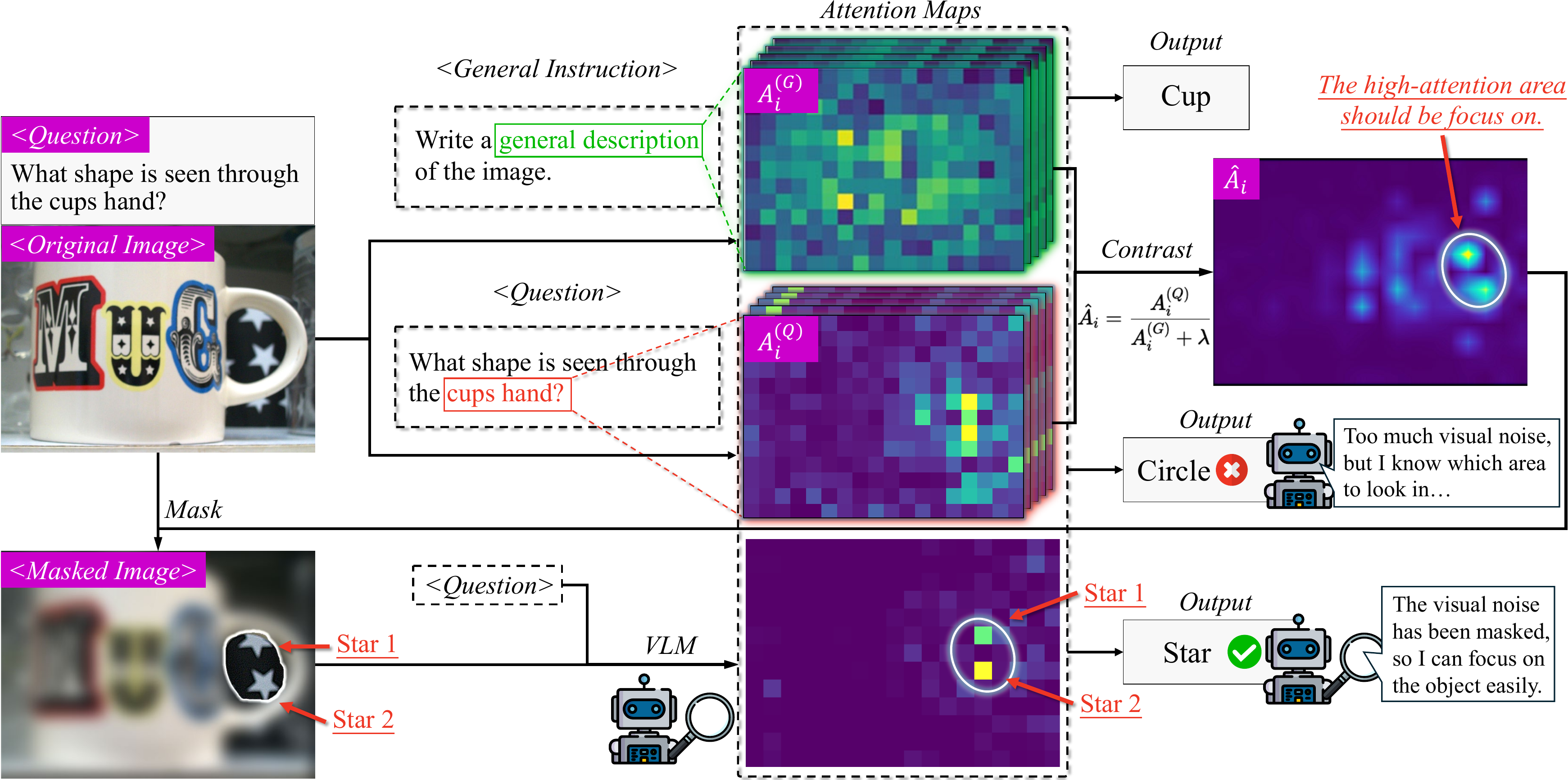}
    \caption{CARVE comprises three stages: Stage 1 generates general attention $A_{i}^{(G)}$ via task-agnostic instructions; Stage 2 extracts task-specific attention $A_{i}^{(Q)}$; Stage 3 applies contrasted attention $\hat{A}_{i}$ to generate enhanced masked images for noise suppression.}
    \label{fig:carve_workflow}
\end{figure}

The general instruction $G$ is a deliberate design choice rather than an arbitrary one: it must elicit global image scanning without focusing on any specific semantic region. Among five candidate prompts, we adopt ``Write a general description of the image.'' as it attains the highest accuracy and the lowest variance while keeping generation short.
When using such general instructions $G$, lacking task-specific semantic cues, the semantic signal function reduces to an approximately uniform distribution ($\mathcal{F}_{\text{sem}}(G, \mathcal{I}) \approx \mathbf{1}_{N_v}$), as illustrated by the diffuse attention pattern in Stage~1 of Fig.~\ref{fig:carve_workflow}.
Quantitatively, the KL divergence from uniform is $\text{KL}(A^{(G)} \| \mathcal{U}) {=} 0.147$, far below the task-specific counterpart $\text{KL}(A^{(Q)} \| \mathcal{U}) {=} 0.812$, confirming that general-instruction attention closely approximates uniformity.
General instruction attention therefore predominantly captures visual noise:
\begin{equation}
A_{l,t}^{(G)}(\mathcal{I}) \approx \mathcal{F}_{\text{vis}}(\mathcal{I}) \otimes \mathbf{1}_{N_v} = \mathcal{F}_{\text{vis}}(\mathcal{I})
\end{equation}

\textbf{Definition 2 (Semantic Extraction Based on Attention Decomposition):} We estimate $\mathcal{F}_{\text{sem}}(Q, \mathcal{I})$ by defining $\hat{A} \in \mathbb{R}^{N_v}_+$ as the solution to:

\begin{equation}
\hat{A} = \arg\min_{\tilde{A} \in \mathcal{A}} \mathcal{J}(\tilde{A}; A^{(Q)}, A^{(G)})
\end{equation}
where the objective function is constructed based on Assumption 1's decomposition:

\begin{equation}
\resizebox{.91\textwidth}{!}{$\displaystyle
\mathcal{J}(\tilde{A}) = \underbrace{\sum_{i=1}^{N_v} \left( \tilde{A}_i \cdot \mathcal{F}_{\text{vis},i}(\mathcal{I}) - [\mathcal{F}_{\text{vis},i}(\mathcal{I}) \cdot \mathcal{F}_{\text{sem},i}(Q,\mathcal{I})] \right)^2}_{\text{Semantic reconstruction error}} + \underbrace{\lambda \sum_{i=1}^{N_v} \tilde{A}_i^2 \cdot \mathcal{F}_{\text{vis},i}(\mathcal{I})}_{\text{Visual suppression regularization}}
$}
\end{equation}

\textbf{Proposition 1 (Closed-form Solution for Semantic Extraction):} Substituting Assumption 1's relationships $A^{(Q)}_i \approx \mathcal{F}_{\text{vis},i} \cdot \mathcal{F}_{\text{sem},i}$ and $A^{(G)}_i \approx \mathcal{F}_{\text{vis},i}$ into the optimization objective yields:

\begin{equation}
\mathcal{J}(\tilde{A}) = \sum_{i=1}^{N_v} \left( \tilde{A}_i \cdot A^{(G)}_i - A^{(Q)}_i \right)^2 + \lambda \sum_{i=1}^{N_v} \tilde{A}_i^2 \cdot A^{(G)}_i
\end{equation}

where $\lambda > 0$ is a regularization parameter that controls the strength of visual noise suppression.

Solving the first-order optimality conditions yields the closed-form solution:

\begin{equation}
\label{eq:contrast}
\hat{A}_i = \frac{A^{(Q)}_i}{A^{(G)}_i + \lambda} = \frac{\mathcal{F}_{\text{vis},i} \cdot \mathcal{F}_{\text{sem},i}}{\mathcal{F}_{\text{vis},i} + \lambda} \approx \mathcal{F}_{\text{sem},i} \quad \text{when } \mathcal{F}_{\text{vis},i} \gg \lambda
\end{equation}

Eq.~\ref{eq:contrast} demonstrates that normalization suppresses the influence of $\mathcal{F}_{\text{vis},i}$ when it dominates (\ie, $\mathcal{F}_{\text{vis},i} \gg \lambda$), approximating the semantic signal $\mathcal{F}_{\text{sem},i}$.

\subsection{Contrastive Attention-based Visual Enhancement}
\label{subsec:visual_enhancement}

Given the refined attention maps $\{\hat{A}\}$, as shown in Algorithm~\ref{alg:carve}, we generate attention masks that physically remove visual noise from the input image.

\paragraph{Attention Maps Fusion.}
Since different layers and time steps capture complementary information, we fuse attention maps across the layer range $\mathcal{L}$ and generation time steps $\mathcal{T} = [t_{\text{start}}, t_{\text{end}}]$ through weighted aggregation.
Later tokens receive higher fusion weights, as they encode richer context from longer preceding sequences.

\paragraph{Mask Generation and Visual Extraction.}
Task-relevant regions are identified by applying the top-p percentile threshold $\tau = \mathcal{Q}_p(S)$, which retains the top $p \in (0,1]$ proportion of pixels from attention map $S$. Connected component analysis extracts coherent regions from the thresholded map.
We select the top-$K$ regions ranked by cumulative attention scores and generate the enhanced image through $\mathcal{I}_{\text{refined}} = \Phi(\mathcal{I}, M^*)$, where $\Phi$ applies masking, cropping, and resizing, and $K$ controls the maximum number of regions to preserve.

\begin{algorithm}[t]
    \caption{\small CARVE: Contrastive Attention Refinement for Visual Enhancement}
    \label{alg:carve}
    \textbf{Notation:} $\mathcal{M}$: VLM model; $\Xi$: attention extraction; $\pi_{H \times W}$: spatial reshape;$\mathcal{Q}_p$: top-$p$ threshold; $\Phi$: visual extraction (mask, crop, resize); $G$: general instruction; $\tau$: threshold; $\mathcal{R}$: connected regions; $K$: max regions to keep
    \begin{algorithmic}[1]
    \REQUIRE $\mathcal{I} \in \mathbb{R}^{H \times W \times 3}$, $Q$, $\mathcal{M}$, $\Theta = \{\mathcal{L}, \mathcal{T}, p, \lambda, K\}$
    \STATE \textbf{Inference:} $\mathcal{A}^Q \leftarrow \{A_{l,t}^{(Q)}\}_{l \in \mathcal{L}, t \in \mathcal{T}} = \Xi(\mathcal{M}, \mathcal{I}, Q)$ \hfill $\triangleright$ Question-specific attention
    \STATE \textbf{Inference:} $\mathcal{A}^G \leftarrow \{A_{l,t}^{(G)}\}_{l \in \mathcal{L}, t \in \mathcal{T}} = \Xi(\mathcal{M}, \mathcal{I}, G)$ \hfill $\triangleright$ General attention
    \STATE \textbf{Contrast:} $\hat{A}_{l,t} \leftarrow \frac{A_{l,t}^{(Q)}}{A_{l,t}^{(G)} + \lambda}$ for all $l \in \mathcal{L}, t \in \mathcal{T}$ \hfill $\triangleright$ following Eq.~\ref{eq:contrast}
    \STATE \textbf{Fuse:} $S \leftarrow \sum_{t \in \mathcal{T}} w_t \sum_{l \in \mathcal{L}} \pi_{H \times W}(\hat{A}_{l,t})$, $w_t = t - t_{\text{start}} + 1$ \hfill $\triangleright$ Weighted attention fusion
    \STATE \textbf{Threshold:} $\tau \leftarrow \mathcal{Q}_p(S)$ \hfill $\triangleright$ Compute threshold to retain top $p$ percentile
    \STATE \textbf{Mask:} $M^* = \bigcup_{k=1}^{K} R_k^*$ where $R_k^* = \arg\max_{R \in \mathcal{R}} \sum_{(i,j) \in R} S(i,j)$ with $\mathcal{R}$ from $S \geq \tau$
    \STATE \textbf{Extract:} $\mathcal{I}_{\text{refined}} \leftarrow \Phi(\mathcal{I}, M^*)$ \hfill $\triangleright$ Visual extraction
    \STATE \textbf{Inference:} \textbf{return} $\mathcal{M}(\mathcal{I}_{\text{refined}}, Q)$ \hfill $\triangleright$ Final inference
    \end{algorithmic}
\end{algorithm}

As shown in Fig.~\ref{fig:carve_workflow}, we propose \textbf{C}ontrastive \textbf{A}ttention \textbf{R}efinement for \textbf{V}isual \textbf{E}nhancement (\textbf{CARVE}), a method that contrasts attention maps to distinguish semantic pixels from noise, preserving only task-relevant regions for enhanced model focus.

\begin{table}[t]
    \centering
    \caption{Accuracy of CARVE across VLMs on four datasets.
    We evaluate three temporal configurations: $t_{\text{start}}$ uses attention from initial generated tokens, $t_{\text{end}}$ from final tokens, and $\mathcal{T}_{\text{full}}$ applies weighted fusion across all tokens.
    We use layer range $\mathcal{L} = [20, 25]$ for attention fusion.
    \textbf{Bold}: best; \underline{underline}: second-best.
    }
    \label{tab:results_control}
    \renewcommand{\arraystretch}{1}
    \resizebox{\textwidth}{!}{
    \setlength{\tabcolsep}{4pt}
    \begin{tabular}{llcccc}
        \toprule
        \textbf{Model} & \textbf{\textbf{Step:} $\mathcal{T}$} & \textbf{A-OKVQA} & \textbf{POPE} & \textbf{V$^*$} & \textbf{TextVQA} \\
        \midrule
        \multirow{4}{*}{\qwenTB}
        & \cellcolor{gray!20}Vanilla & \cellcolor{gray!20}73.0 & \cellcolor{gray!20}86.9 & \cellcolor{gray!20}50.3 & \cellcolor{gray!20}72.8 \\
        &  $t_{\text{start}}$ & 76.5$_{\scriptscriptstyle\blacktriangle\,4.79}$ & 87.1$_{\scriptscriptstyle\blacktriangle\,0.23}$ & 56.0$_{\scriptscriptstyle\blacktriangle\,11.33}$ & 76.1$_{\scriptscriptstyle\blacktriangle\,4.53}$ \\
        &  $t_{\text{end}}$ & \goodmetric{79.2}$_{{\color{ggreen}\scriptscriptstyle\blacktriangle}\,8.49}$ & \goodmetric{88.4}$_{{\color{ggreen}\scriptscriptstyle\blacktriangle}\,1.73}$ & \goodmetric{57.1}$_{{\color{ggreen}\scriptscriptstyle\blacktriangle}\,13.52}$ & \goodmetric{76.4}$_{{\color{ggreen}\scriptscriptstyle\blacktriangle}\,4.95}$ \\
        &  $\mathcal{T}_{\text{full}}$ & \secondmetric{78.3}$_{\scriptscriptstyle\blacktriangle\,7.26}$ & \secondmetric{87.9}$_{\scriptscriptstyle\blacktriangle\,1.15}$ & \secondmetric{56.5}$_{\scriptscriptstyle\blacktriangle\,12.33}$ & \secondmetric{76.3}$_{\scriptscriptstyle\blacktriangle\,4.81}$ \\
        \cmidrule(lr){1-6}
        \multirow{4}{*}{\qwenSB}
        & \cellcolor{gray!20}Vanilla & \cellcolor{gray!20}75.0 & \cellcolor{gray!20}87.0 & \cellcolor{gray!20}50.8 & \cellcolor{gray!20}75.0 \\
        &  $t_{\text{start}}$ & 77.0$_{\scriptscriptstyle\blacktriangle\,2.67}$ & 87.9$_{\scriptscriptstyle\blacktriangle\,1.03}$ & \secondmetric{58.6}$_{\scriptscriptstyle\blacktriangle\,15.35}$ & 80.7$_{\scriptscriptstyle\blacktriangle\,7.60}$ \\
        &  $t_{\text{end}}$ & \goodmetric{78.3}$_{{\color{ggreen}\scriptscriptstyle\blacktriangle}\,4.40}$ & \goodmetric{89.7}$_{{\color{ggreen}\scriptscriptstyle\blacktriangle}\,3.10}$ & \goodmetric{59.7}$_{{\color{ggreen}\scriptscriptstyle\blacktriangle}\,17.52}$ & \goodmetric{81.9}$_{{\color{ggreen}\scriptscriptstyle\blacktriangle}\,9.20}$ \\
        &  $\mathcal{T}_{\text{full}}$ & \secondmetric{78.0}$_{\scriptscriptstyle\blacktriangle\,4.00}$ & \secondmetric{88.6}$_{\scriptscriptstyle\blacktriangle\,1.84}$ & 58.1$_{\scriptscriptstyle\blacktriangle\,14.37}$ & \secondmetric{81.7}$_{\scriptscriptstyle\blacktriangle\,8.93}$ \\
        \cmidrule(lr){1-6}
        \multirow{4}{*}{\llavaSB}
        & \cellcolor{gray!20}Vanilla & \cellcolor{gray!20}71.5 & \cellcolor{gray!20}83.6 & \cellcolor{gray!20}38.7 & \cellcolor{gray!20}47.8 \\
        &  $t_{\text{start}}$ & 73.9$_{\scriptscriptstyle\blacktriangle\,3.36}$ & \secondmetric{86.8}$_{\scriptscriptstyle\blacktriangle\,3.83}$ & \secondmetric{57.1}$_{\scriptscriptstyle\blacktriangle\,47.55}$ & \secondmetric{57.9}$_{\scriptscriptstyle\blacktriangle\,21.13}$ \\
        &  $t_{\text{end}}$ & \goodmetric{78.2}$_{{\color{ggreen}\scriptscriptstyle\blacktriangle}\,9.37}$ & \goodmetric{89.0}$_{{\color{ggreen}\scriptscriptstyle\blacktriangle}\,6.46}$ & \goodmetric{66.5}$_{{\color{ggreen}\scriptscriptstyle\blacktriangle}\,71.83}$ & \goodmetric{58.2}$_{{\color{ggreen}\scriptscriptstyle\blacktriangle}\,21.76}$ \\
        &  $\mathcal{T}_{\text{full}}$ & \secondmetric{75.4}$_{\scriptscriptstyle\blacktriangle\,5.45}$ & \goodmetric{89.0}$_{{\color{ggreen}\scriptscriptstyle\blacktriangle}\,6.46}$ & \goodmetric{66.5}$_{{\color{ggreen}\scriptscriptstyle\blacktriangle}\,71.83}$ & \secondmetric{57.9}$_{\scriptscriptstyle\blacktriangle\,21.13}$ \\
        \cmidrule(lr){1-6}
        \multirow{4}{*}{\llavaffB}
        & \cellcolor{gray!20}Vanilla & \cellcolor{gray!20}75.7 & \cellcolor{gray!20}84.6 & \cellcolor{gray!20}42.4 & \cellcolor{gray!20}57.1 \\
        &  $t_{\text{start}}$ & \secondmetric{76.2}$_{\scriptscriptstyle\blacktriangle\,0.66}$ & 90.0$_{\scriptscriptstyle\blacktriangle\,6.38}$ & 65.4$_{\scriptscriptstyle\blacktriangle\,54.25}$ & \secondmetric{59.2}$_{\scriptscriptstyle\blacktriangle\,3.68}$ \\
        &  $t_{\text{end}}$ & \goodmetric{76.9}$_{{\color{ggreen}\scriptscriptstyle\blacktriangle}\,1.59}$ & \goodmetric{90.7}$_{{\color{ggreen}\scriptscriptstyle\blacktriangle}\,7.21}$ & \goodmetric{74.3}$_{{\color{ggreen}\scriptscriptstyle\blacktriangle}\,75.24}$ & \goodmetric{61.2}$_{{\color{ggreen}\scriptscriptstyle\blacktriangle}\,7.18}$ \\
        &  $\mathcal{T}_{\text{full}}$ & \goodmetric{76.9}$_{{\color{ggreen}\scriptscriptstyle\blacktriangle}\,1.59}$ & \secondmetric{90.1}$_{\scriptscriptstyle\blacktriangle\,6.50}$ & \secondmetric{70.2}$_{\scriptscriptstyle\blacktriangle\,65.57}$ & \goodmetric{61.2}$_{{\color{ggreen}\scriptscriptstyle\blacktriangle}\,7.18}$ \\
        \bottomrule
    \end{tabular}
    }
\end{table}

\begin{table}[t]
    \centering
    \caption{
    We investigate CARVE's accuracy using both single-layer and multi-layer intervention strategies at shallow, middle, and deep model depths, where single-layer interventions use attention maps $\hat{A}_i$ from individual layers, while multi-layer interventions fuse maps across multiple layers to guide masking decisions.
    We employ $\mathcal{T}_{\text{full}}$ as the time step configuration.}
    \label{tab:layers_ablation}
    \renewcommand{\arraystretch}{1}
    \resizebox{\textwidth}{!}{
    \setlength{\tabcolsep}{4pt}
    \begin{tabular}{lllcccc}
        \toprule
        \textbf{Model} & \multicolumn{2}{c}{\textbf{Layer(s):} $\mathcal{L}$}  & \textbf{A-OKVQA} & \textbf{POPE} & \textbf{V$^*$} & \textbf{TextVQA} \\
        \midrule
        \multirow{7}{*}{\qwenTB}
        & \multicolumn{2}{c}{\cellcolor{gray!20}Vanilla} & \cellcolor{gray!20}73.0 & \cellcolor{gray!20}86.9 & \cellcolor{gray!20}50.3 & \cellcolor{gray!20}72.8 \\
        & \multirow{3}{*}{Single Layer}
        & 14 & 74.3$_{\scriptscriptstyle\blacktriangle\,1.78}$ & 87.1$_{\scriptscriptstyle\blacktriangle\,0.23}$ & 53.9$_{\scriptscriptstyle\blacktriangle\,7.16}$ & 73.6$_{\scriptscriptstyle\blacktriangle\,1.10}$ \\
        & & 20 & 76.5$_{\scriptscriptstyle\blacktriangle\,4.79}$ & 87.4$_{\scriptscriptstyle\blacktriangle\,0.58}$ & \secondmetric{56.0}$_{\scriptscriptstyle\blacktriangle\,11.33}$ & 74.7$_{\scriptscriptstyle\blacktriangle\,2.61}$ \\
        & & 25 & 76.7$_{\scriptscriptstyle\blacktriangle\,5.07}$ & 87.5$_{\scriptscriptstyle\blacktriangle\,0.69}$ & \secondmetric{56.0}$_{\scriptscriptstyle\blacktriangle\,11.33}$ & 75.9$_{\scriptscriptstyle\blacktriangle\,4.26}$ \\
        \cmidrule{2-7}
        & \multirow{3}{*}{Multi-Layers}
        & [10, 15] & 74.0$_{\scriptscriptstyle\blacktriangle\,1.37}$ & 86.9$_{0.00}$ & 53.4$_{\scriptscriptstyle\blacktriangle\,6.16}$ & 73.0$_{\scriptscriptstyle\blacktriangle\,0.27}$ \\
        & & [15, 20] & \secondmetric{76.8}$_{\scriptscriptstyle\blacktriangle\,5.21}$ & \secondmetric{87.7}$_{\scriptscriptstyle\blacktriangle\,0.92}$ & \secondmetric{56.0}$_{\scriptscriptstyle\blacktriangle\,11.33}$ & \secondmetric{76.0}$_{\scriptscriptstyle\blacktriangle\,4.40}$ \\
        & & [20, 25] & \goodmetric{78.3}$_{{\color{ggreen}\scriptscriptstyle\blacktriangle}\,7.26}$ & \goodmetric{87.9}$_{{\color{ggreen}\scriptscriptstyle\blacktriangle}\,1.15}$ & \goodmetric{56.5}$_{{\color{ggreen}\scriptscriptstyle\blacktriangle}\,12.33}$ & \goodmetric{76.3}$_{{\color{ggreen}\scriptscriptstyle\blacktriangle}\,4.81}$ \\
        \cmidrule(lr){1-7}
        \multirow{7}{*}{\qwenSB}
        & \multicolumn{2}{c}{\cellcolor{gray!20}Vanilla} & \cellcolor{gray!20}75.0 & \cellcolor{gray!20}87.0 & \cellcolor{gray!20}50.8 & \cellcolor{gray!20}75.0 \\
        & \multirow{3}{*}{Single Layer}
        & 14 & 75.2$_{\scriptscriptstyle\blacktriangle\,0.27}$ & 87.5$_{\scriptscriptstyle\blacktriangle\,0.57}$ & 54.5$_{\scriptscriptstyle\blacktriangle\,7.28}$ & 75.2$_{\scriptscriptstyle\blacktriangle\,0.27}$ \\
        & & 20 & 76.9$_{\scriptscriptstyle\blacktriangle\,2.53}$ & 87.9$_{\scriptscriptstyle\blacktriangle\,1.03}$ & 56.5$_{\scriptscriptstyle\blacktriangle\,11.22}$ & 77.9$_{\scriptscriptstyle\blacktriangle\,3.87}$ \\
        & & 25 & 77.0$_{\scriptscriptstyle\blacktriangle\,2.67}$ & 88.2$_{\scriptscriptstyle\blacktriangle\,1.38}$ & 57.1$_{\scriptscriptstyle\blacktriangle\,12.40}$ & 78.4$_{\scriptscriptstyle\blacktriangle\,4.53}$ \\
        \cmidrule{2-7}
        & \multirow{3}{*}{Multi-Layers}
        & [10, 15] & 75.0$_{0.00}$ & 87.0$_{0.00}$ & 51.3$_{\scriptscriptstyle\blacktriangle\,0.98}$ & 75.0$_{0.00}$ \\
        & & [15, 20] & \secondmetric{77.1}$_{\scriptscriptstyle\blacktriangle\,2.80}$ & \secondmetric{88.4}$_{\scriptscriptstyle\blacktriangle\,1.61}$ & \secondmetric{57.6}$_{\scriptscriptstyle\blacktriangle\,13.39}$ & \secondmetric{79.5}$_{\scriptscriptstyle\blacktriangle\,6.00}$ \\
        & & [20, 25] & \goodmetric{78.0}$_{{\color{ggreen}\scriptscriptstyle\blacktriangle}\,4.00}$ & \goodmetric{88.6}$_{{\color{ggreen}\scriptscriptstyle\blacktriangle}\,1.84}$ & \goodmetric{58.1}$_{{\color{ggreen}\scriptscriptstyle\blacktriangle}\,14.37}$ & \goodmetric{81.7}$_{{\color{ggreen}\scriptscriptstyle\blacktriangle}\,8.93}$ \\
        \cmidrule(lr){1-7}
        \multirow{7}{*}{\llavaSB}
        & \multicolumn{2}{c}{\cellcolor{gray!20}Vanilla} & \cellcolor{gray!20}71.5 & \cellcolor{gray!20}83.6 & \cellcolor{gray!20}38.7 & \cellcolor{gray!20}47.8 \\
        & \multirow{3}{*}{Single Layer}
        & 14 & 71.7$_{\scriptscriptstyle\blacktriangle\,0.28}$ & 85.1$_{\scriptscriptstyle\blacktriangle\,1.79}$ & 63.4$_{\scriptscriptstyle\blacktriangle\,63.82}$ & 54.0$_{\scriptscriptstyle\blacktriangle\,12.97}$ \\
        & & 20 & 74.0$_{\scriptscriptstyle\blacktriangle\,3.50}$ & 87.2$_{\scriptscriptstyle\blacktriangle\,4.31}$ & \secondmetric{65.4}$_{\scriptscriptstyle\blacktriangle\,68.99}$ & 56.1$_{\scriptscriptstyle\blacktriangle\,17.36}$ \\
        & & 25 & 74.1$_{\scriptscriptstyle\blacktriangle\,3.64}$ & 87.1$_{\scriptscriptstyle\blacktriangle\,4.19}$ & \secondmetric{65.4}$_{\scriptscriptstyle\blacktriangle\,68.99}$ & 56.2$_{\scriptscriptstyle\blacktriangle\,17.57}$ \\
        \cmidrule{2-7}
        & \multirow{3}{*}{Multi-Layers}
        & [10, 15] & 71.5$_{0.00}$ & 84.5$_{\scriptscriptstyle\blacktriangle\,1.08}$ &48.2$_{\scriptscriptstyle\blacktriangle\,24.55}$ & 49.2$_{\scriptscriptstyle\blacktriangle\,2.93}$ \\
        & & [15, 20] & \secondmetric{74.2}$_{\scriptscriptstyle\blacktriangle\,3.78}$ & \secondmetric{87.5}$_{\scriptscriptstyle\blacktriangle\,4.67}$ & \secondmetric{65.4}$_{\scriptscriptstyle\blacktriangle\,68.99}$ & \secondmetric{56.4}$_{\scriptscriptstyle\blacktriangle\,17.99}$ \\
        & & [20, 25] & \goodmetric{75.4}$_{{\color{ggreen}\scriptscriptstyle\blacktriangle}\,5.45}$ & \goodmetric{89.0}$_{{\color{ggreen}\scriptscriptstyle\blacktriangle}\,6.46}$ & \goodmetric{66.5}$_{{\color{ggreen}\scriptscriptstyle\blacktriangle}\,71.83}$ & \goodmetric{57.9}$_{{\color{ggreen}\scriptscriptstyle\blacktriangle}\,21.13}$ \\
        \cmidrule(lr){1-7}
        \multirow{7}{*}{\llavaffB}
        & \multicolumn{2}{c}{\cellcolor{gray!20}Vanilla} & \cellcolor{gray!20}75.7 & \cellcolor{gray!20}84.6 & \cellcolor{gray!20}42.4 & \cellcolor{gray!20}57.1 \\
        & \multirow{3}{*}{Single Layer}
        & 14 & 75.8$_{\scriptscriptstyle\blacktriangle\,0.13}$ & 86.1$_{\scriptscriptstyle\blacktriangle\,1.77}$ & 66.5$_{\scriptscriptstyle\blacktriangle\,56.84}$ & 58.2$_{\scriptscriptstyle\blacktriangle\,1.93}$ \\
        & & 20 & 76.2$_{\scriptscriptstyle\blacktriangle\,0.66}$ & 88.2$_{\scriptscriptstyle\blacktriangle\,4.26}$ & 68.6$_{\scriptscriptstyle\blacktriangle\,61.79}$ & 59.1$_{\scriptscriptstyle\blacktriangle\,3.50}$ \\
        & & 25 & 76.2$_{\scriptscriptstyle\blacktriangle\,0.66}$ & 88.1$_{\scriptscriptstyle\blacktriangle\,4.14}$ & \secondmetric{69.1}$_{\scriptscriptstyle\blacktriangle\,62.97}$ & 59.2$_{\scriptscriptstyle\blacktriangle\,3.68}$ \\
        \cmidrule{2-7}
        & \multirow{3}{*}{Multi-Layers}
        & [10, 15] & 75.7$_{0.00}$ & 85.0$_{\scriptscriptstyle\blacktriangle\,0.47}$ & 52.9$_{\scriptscriptstyle\blacktriangle\,24.76}$ & 57.4$_{\scriptscriptstyle\blacktriangle\,0.53}$ \\
        & & [15, 20] & \secondmetric{76.8}$_{\scriptscriptstyle\blacktriangle\,1.45}$ & \secondmetric{88.6}$_{\scriptscriptstyle\blacktriangle\,4.73}$ & \secondmetric{69.1}$_{\scriptscriptstyle\blacktriangle\,62.97}$ & \secondmetric{59.4}$_{\scriptscriptstyle\blacktriangle\,4.03}$ \\
        & & [20, 25] & \goodmetric{76.9}$_{{\color{ggreen}\scriptscriptstyle\blacktriangle}\,1.59}$ & \goodmetric{90.1}$_{{\color{ggreen}\scriptscriptstyle\blacktriangle}\,6.50}$ & \goodmetric{70.2}$_{{\color{ggreen}\scriptscriptstyle\blacktriangle}\,65.57}$ & \goodmetric{61.2}$_{{\color{ggreen}\scriptscriptstyle\blacktriangle}\,7.18}$ \\
        \bottomrule
    \end{tabular}
    }
\end{table}

\begin{table}[t]
    \centering
    \caption{Extended cross-dataset evaluation on three additional benchmarks. DocVQA and ChartQA primarily involve local visual perception, while ScienceQA requires knowledge-based reasoning.}
    \label{tab:extended_datasets}
    \renewcommand{\arraystretch}{1}
    \resizebox{\textwidth}{!}{
    \setlength{\tabcolsep}{2pt}
    \begin{tabular}{llccccc}
        \toprule
        \multirow{2}{*}{\textbf{Model}} & & \multicolumn{2}{c}{\textbf{DocVQA}} & \multicolumn{2}{c}{\textbf{ChartQA}} & \textbf{ScienceQA} \\
        & & \textbf{EM} & \textbf{ANLS} & \textbf{EM} & \textbf{Relaxed} & \textbf{MC Acc} \\
        \midrule
        \multirow{4}{*}{\qwenTB}
        & \cellcolor{gray!20}Vanilla & \cellcolor{gray!20}49.70 & \cellcolor{gray!20}66.07 & \cellcolor{gray!20}67.64 & \cellcolor{gray!20}77.92 & \cellcolor{gray!20}75.11 \\
        & ViCrop & 55.80$_{\scriptscriptstyle\blacktriangle\,12.27}$ & 72.27$_{\scriptscriptstyle\blacktriangle\,9.38}$ & 68.92$_{\scriptscriptstyle\blacktriangle\,1.89}$ & 79.64$_{\scriptscriptstyle\blacktriangle\,2.21}$ & 75.21$_{\scriptscriptstyle\blacktriangle\,0.13}$ \\
        & SD-RPN & \secondmetric{60.90}$_{\scriptscriptstyle\blacktriangle\,22.54}$ & \secondmetric{73.50}$_{\scriptscriptstyle\blacktriangle\,11.25}$ & \secondmetric{69.88}$_{\scriptscriptstyle\blacktriangle\,3.31}$ & \secondmetric{80.56}$_{\scriptscriptstyle\blacktriangle\,3.39}$ & \secondmetric{75.36}$_{\scriptscriptstyle\blacktriangle\,0.33}$ \\
        & \cellcolor{ours}CARVE & \cellcolor{ours}\goodmetric{61.70}$_{{\color{ggreen}\scriptscriptstyle\blacktriangle}\,24.14}$ & \cellcolor{ours}\goodmetric{74.94}$_{{\color{ggreen}\scriptscriptstyle\blacktriangle}\,13.43}$ & \cellcolor{ours}\goodmetric{70.80}$_{{\color{ggreen}\scriptscriptstyle\blacktriangle}\,4.67}$ & \cellcolor{ours}\goodmetric{81.76}$_{{\color{ggreen}\scriptscriptstyle\blacktriangle}\,4.93}$ & \cellcolor{ours}\goodmetric{75.56}$_{{\color{ggreen}\scriptscriptstyle\blacktriangle}\,0.60}$ \\
        \cmidrule(lr){1-7}
        \multirow{4}{*}{\qwenSB}
        & \cellcolor{gray!20}Vanilla & \cellcolor{gray!20}54.30 & \cellcolor{gray!20}69.97 & \cellcolor{gray!20}69.32 & \cellcolor{gray!20}79.12 & \cellcolor{gray!20}81.21 \\
        & ViCrop & \secondmetric{57.60}$_{\scriptscriptstyle\blacktriangle\,6.08}$ & 70.84$_{\scriptscriptstyle\blacktriangle\,1.24}$ & 69.76$_{\scriptscriptstyle\blacktriangle\,0.63}$ & 79.48$_{\scriptscriptstyle\blacktriangle\,0.46}$ & 81.56$_{\scriptscriptstyle\blacktriangle\,0.43}$ \\
        & SD-RPN & 55.90$_{\scriptscriptstyle\blacktriangle\,2.95}$ & \secondmetric{72.23}$_{\scriptscriptstyle\blacktriangle\,3.23}$ & \secondmetric{70.36}$_{\scriptscriptstyle\blacktriangle\,1.50}$ & \goodmetric{80.12}$_{{\color{ggreen}\scriptscriptstyle\blacktriangle}\,1.26}$ & \goodmetric{83.84}$_{{\color{ggreen}\scriptscriptstyle\blacktriangle}\,3.24}$ \\
        & \cellcolor{ours}CARVE & \cellcolor{ours}\goodmetric{63.00}$_{{\color{ggreen}\scriptscriptstyle\blacktriangle}\,16.02}$ & \cellcolor{ours}\goodmetric{75.95}$_{{\color{ggreen}\scriptscriptstyle\blacktriangle}\,8.55}$ & \cellcolor{ours}\goodmetric{72.48}$_{{\color{ggreen}\scriptscriptstyle\blacktriangle}\,4.56}$ & \cellcolor{ours}\secondmetric{79.96}$_{\scriptscriptstyle\blacktriangle\,1.06}$ & \cellcolor{ours}\secondmetric{82.50}$_{\scriptscriptstyle\blacktriangle\,1.59}$ \\
        \cmidrule(lr){1-7}
        \multirow{4}{*}{\llavaSB}
        & \cellcolor{gray!20}Vanilla & \cellcolor{gray!20}11.20 & \cellcolor{gray!20}19.78 & \cellcolor{gray!20}8.96 & \cellcolor{gray!20}17.88 & \cellcolor{gray!20}59.10 \\
        & ViCrop & 11.90$_{\scriptscriptstyle\blacktriangle\,6.25}$ & 27.00$_{\scriptscriptstyle\blacktriangle\,36.50}$ & 9.12$_{\scriptscriptstyle\blacktriangle\,1.79}$ & 18.12$_{\scriptscriptstyle\blacktriangle\,1.34}$ & 59.49$_{\scriptscriptstyle\blacktriangle\,0.66}$ \\
        & SD-RPN & \secondmetric{12.20}$_{\scriptscriptstyle\blacktriangle\,8.93}$ & \goodmetric{29.20}$_{{\color{ggreen}\scriptscriptstyle\blacktriangle}\,47.62}$ & \secondmetric{10.72}$_{\scriptscriptstyle\blacktriangle\,19.64}$ & \secondmetric{20.12}$_{\scriptscriptstyle\blacktriangle\,12.53}$ & \secondmetric{59.84}$_{\scriptscriptstyle\blacktriangle\,1.25}$ \\
        & \cellcolor{ours}CARVE & \cellcolor{ours}\goodmetric{12.40}$_{{\color{ggreen}\scriptscriptstyle\blacktriangle}\,10.71}$ & \cellcolor{ours}\secondmetric{28.21}$_{\scriptscriptstyle\blacktriangle\,42.62}$ & \cellcolor{ours}\goodmetric{11.20}$_{{\color{ggreen}\scriptscriptstyle\blacktriangle}\,25.00}$ & \cellcolor{ours}\goodmetric{20.60}$_{{\color{ggreen}\scriptscriptstyle\blacktriangle}\,15.21}$ & \cellcolor{ours}\goodmetric{60.39}$_{{\color{ggreen}\scriptscriptstyle\blacktriangle}\,2.18}$ \\
        \cmidrule(lr){1-7}
        \multirow{4}{*}{\llavaffB}
        & \cellcolor{gray!20}Vanilla & \cellcolor{gray!20}14.00 & \cellcolor{gray!20}24.42 & \cellcolor{gray!20}9.88 & \cellcolor{gray!20}18.20 & \cellcolor{gray!20}61.68 \\
        & ViCrop & 14.40$_{\scriptscriptstyle\blacktriangle\,2.86}$ & 31.20$_{\scriptscriptstyle\blacktriangle\,27.76}$ & \secondmetric{12.08}$_{\scriptscriptstyle\blacktriangle\,22.27}$ & \secondmetric{20.12}$_{\scriptscriptstyle\blacktriangle\,10.55}$ & 62.92$_{\scriptscriptstyle\blacktriangle\,2.01}$ \\
        & SD-RPN & \secondmetric{15.90}$_{\scriptscriptstyle\blacktriangle\,13.57}$ & \secondmetric{33.32}$_{\scriptscriptstyle\blacktriangle\,36.45}$ & 11.64$_{\scriptscriptstyle\blacktriangle\,17.81}$ & 19.76$_{\scriptscriptstyle\blacktriangle\,8.57}$ & \secondmetric{63.46}$_{\scriptscriptstyle\blacktriangle\,2.89}$ \\
        & \cellcolor{ours}CARVE & \cellcolor{ours}\goodmetric{16.20}$_{{\color{ggreen}\scriptscriptstyle\blacktriangle}\,15.71}$ & \cellcolor{ours}\goodmetric{33.48}$_{{\color{ggreen}\scriptscriptstyle\blacktriangle}\,37.10}$ & \cellcolor{ours}\goodmetric{13.32}$_{{\color{ggreen}\scriptscriptstyle\blacktriangle}\,34.82}$ & \cellcolor{ours}\goodmetric{21.20}$_{{\color{ggreen}\scriptscriptstyle\blacktriangle}\,16.48}$ & \cellcolor{ours}\goodmetric{65.34}$_{{\color{ggreen}\scriptscriptstyle\blacktriangle}\,5.93}$ \\
        \bottomrule
    \end{tabular}
    }
\end{table}

\begin{figure}[t]
    \centering
    \begin{subfigure}[b]{0.48\textwidth}
        \includegraphics[width=\textwidth]{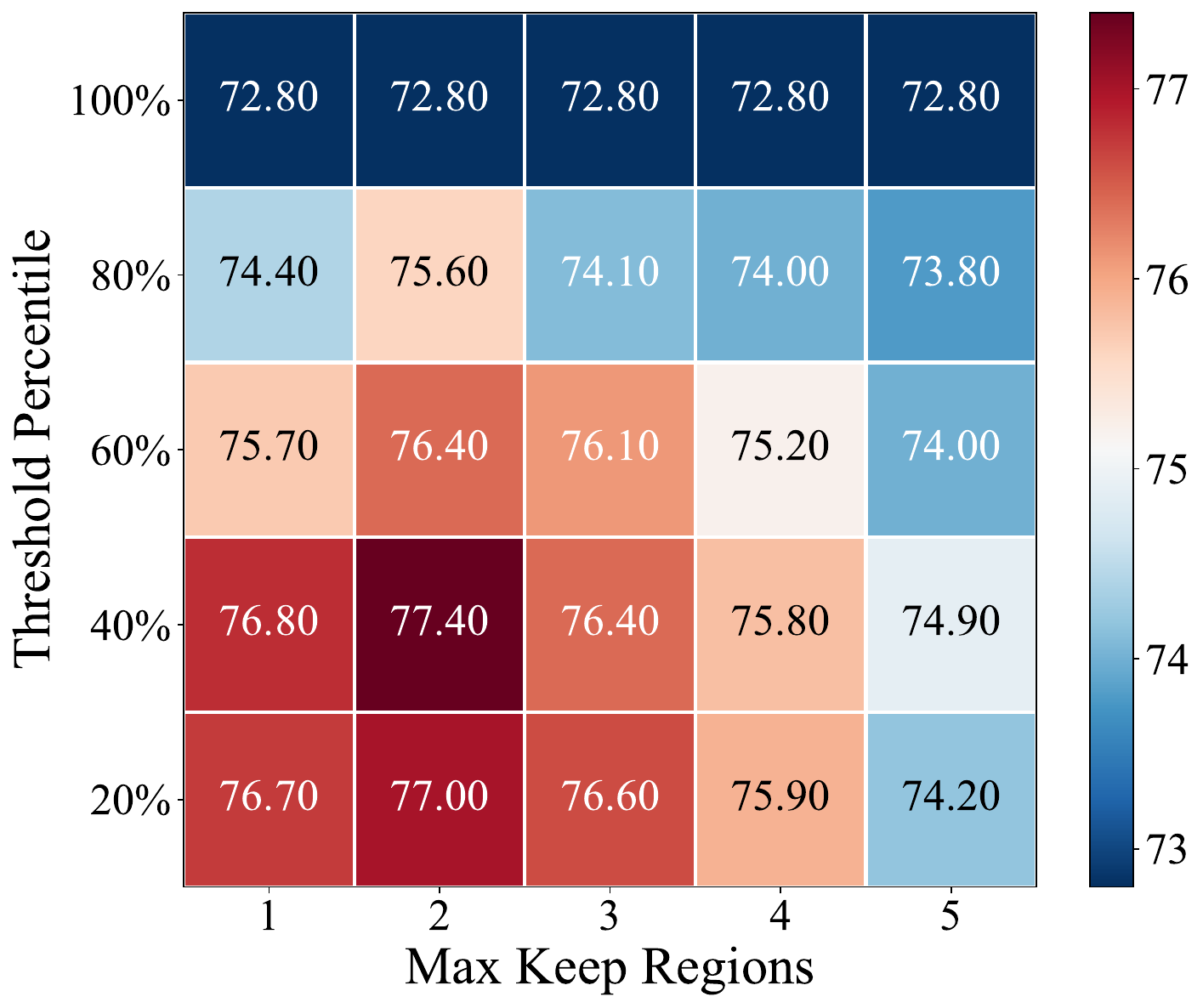}
        \caption{Accuracy for \qwenTB on TextVQA}
        \label{fig:qwen3b_grid_a}
    \end{subfigure}
    \hfill
    \begin{subfigure}[b]{0.48\textwidth}
        \includegraphics[width=\textwidth]{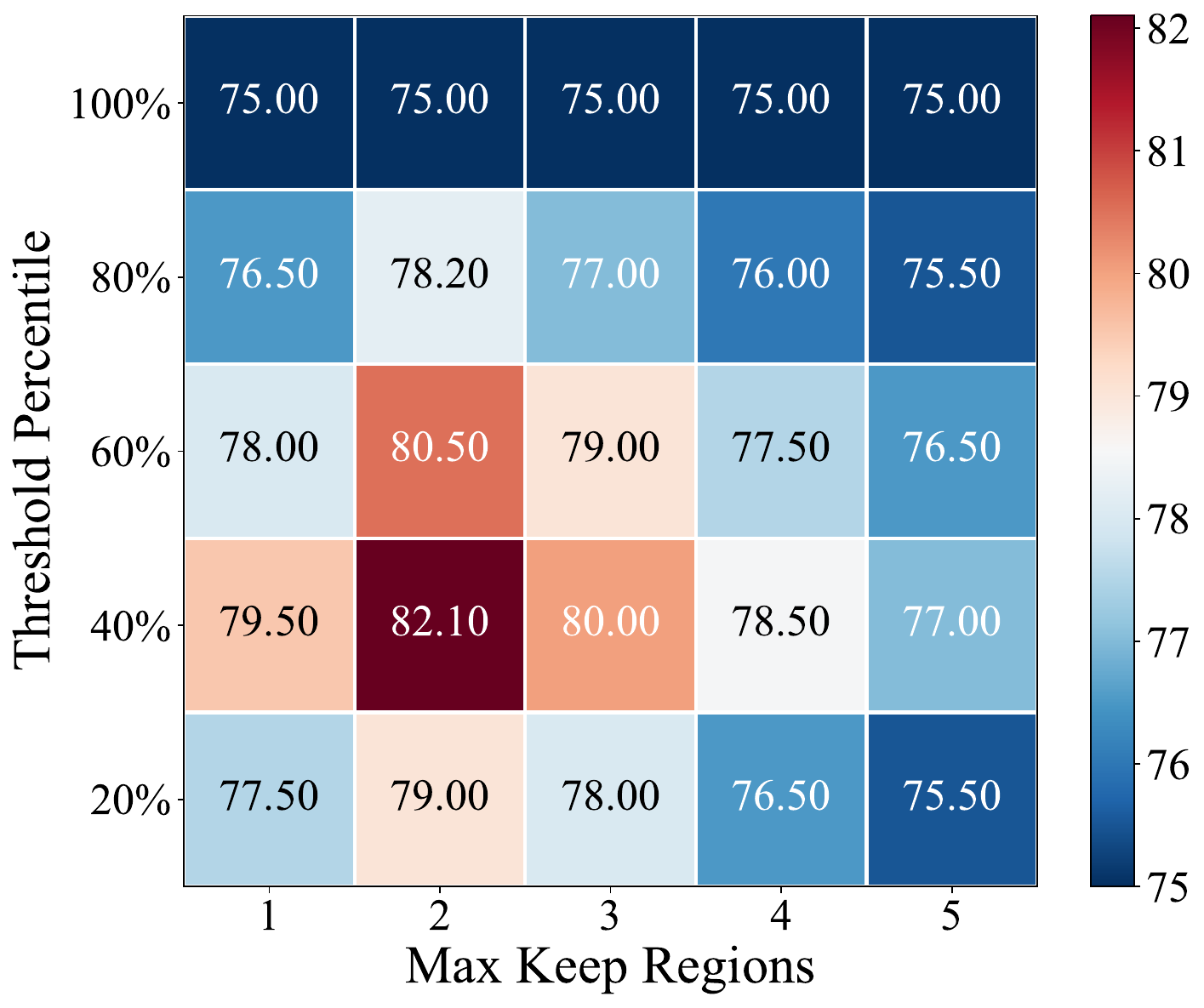}
        \caption{Accuracy for \qwenSB on TextVQA}
        \label{fig:qwen3b_grid_b}
    \end{subfigure}
    \caption{Mask generation hyperparameter impact on TextVQA accuracy for \textsc{Qwen} family, across top-$p$ threshold and maximum keep regions $K$.
    }
    \label{fig:heatmaps}
\end{figure}

\section{Experiments}
\label{sec:results}

\subsection{Experimental Setup}
\label{subsec:experimental_setup}

\paragraph{Datasets.}
We conduct experiments on seven datasets spanning diverse task dimensions.
For visual reasoning and understanding, we use A-OKVQA~\cite{schwenk2022aokvqa}, POPE~\cite{li2023evaluating}, V$^*$~\cite{wu2024v}, and TextVQA~\cite{singh2019vqa}.
To further evaluate generalizability, we extend evaluation to DocVQA~\cite{mathew2021docvqa} for document understanding (Exact Match and ANLS), ChartQA~\cite{masry2022chartqa} for chart comprehension (Exact Match and Relaxed Accuracy), and ScienceQA~\cite{lu2022learn} for knowledge-based reasoning (multiple-choice accuracy).
For TextVQA, we provide only images and questions without external OCR augmentation, following the OCR-free protocol of ViCrop~\cite{zhang2025mllms}.

\paragraph{Models.}
We conduct experiments on four VLMs: \textsc{Qwen2.5-VL-3B-Instruct}, \textsc{Qwen2.5-VL-7B-Instruct}~\cite{bai2025qwen25vl}, \textsc{LLaVA-1.5-7B}, and \textsc{LLaVA-1.5-13B}~\cite{liu2023visual}. The Qwen family processes images at $448 \times 448$ resolution, while the LLaVA-1.5 family operates at $336 \times 336$ resolution. All models employ greedy decoding with regularization parameter $\lambda = 0.05$.

\paragraph{Baselines.}
We compare CARVE against two concurrent methods that also enhance VLM visual perception: ViCrop~\cite{zhang2025mllms} and SD-RPN~\cite{shi2026catching}.

\subsection{Results}

\paragraph{Main Results.}
Tables~\ref{tab:results_control} and~\ref{tab:layers_ablation} show that CARVE improves performance across all evaluated models and datasets, with larger gains on perceptually demanding tasks.
Earlier-generation models show larger gains than their more recent counterparts.
For instance, \llavaSB gains +27.8 absolute points on V$^*$, whereas \qwenSB gains +8.9 points.
This pattern indicates that less capable models suffer more from visual complexity interference and benefit more from contrastive attention focusing.
The particularly large V$^*$ gains are attributable to its benchmark design: every sample requires localizing a fine-grained target within a cluttered scene, precisely the scenario CARVE addresses.
Despite the modest sample size, the improvement on V$^*$ for \llavaSB is statistically significant: the 95\% bootstrap confidence intervals do not overlap, spanning 31.9--46.1\% for the vanilla model versus 59.7--72.8\% for CARVE, and a paired McNemar's test further confirms significance at $p < 0.001$.

\paragraph{Cross-Dataset Comparison.}
Table~\ref{tab:extended_datasets} compares CARVE with ViCrop and SD-RPN on three additional benchmarks, where SD-RPN additionally requires training.
CARVE, despite being entirely training-free, achieves the best results in the majority of settings.
Improvements scale with visual perception demands: DocVQA, requiring fine-grained text recognition in dense layouts, yields the largest relative gains of up to +24.14\% EM for \qwenTB; ChartQA shows moderate improvements of +4.67\% EM for \qwenTB; and ScienceQA, relying on knowledge reasoning, yields consistently positive yet smaller gains ranging from +0.60\% to +5.93\%.
This gradient across tasks suggests that suppressing visual noise benefits tasks roughly in proportion to their perceptual demands, with no observed degradation of reasoning performance.

\paragraph{Time Step Ablation.}
Table~\ref{tab:results_control} shows a clear ranking among time step selection strategies: $t_{\text{end}}$ generally outperforms $\mathcal{T}_{\text{full}}$, which in turn surpasses $t_{\text{start}}$.
For example, on TextVQA with \qwenSB, $t_{\text{end}}$ reaches 81.9\%, $\mathcal{T}_{\text{full}}$ 81.7\%, and $t_{\text{start}}$ 80.7\%.
This ranking is consistent with autoregressive decoding: later tokens access longer contexts, so the final token's attention maps better localize task-relevant objects for CARVE's masking.

\paragraph{Layer Selection Ablation.}
Table~\ref{tab:layers_ablation} reports the effect of layer selection on attention extraction.
Across all models, [20,25] consistently achieves the best performance while [10,15] ranks last.
The general trend from best to worst is:
[20,25], [15,20], single layer 25, single layer 20, single layer 14, and [10,15], with occasional ties among adjacent single-layer configurations.
For instance, \llavaSB on TextVQA gains +21.13\% with [20,25], +17.99\% with [15,20], but only +2.93\% with [10,15].
Multi-layer fusion captures complementary patterns while reducing noise, consistent with the layer-wise entropy trend in Fig.~\ref{fig:entropy_performance}: early layers perform global scanning with high entropy, while deeper layers focus on task-relevant patterns.

\begin{figure}[t]
    \centering
    \includegraphics[width=\linewidth]{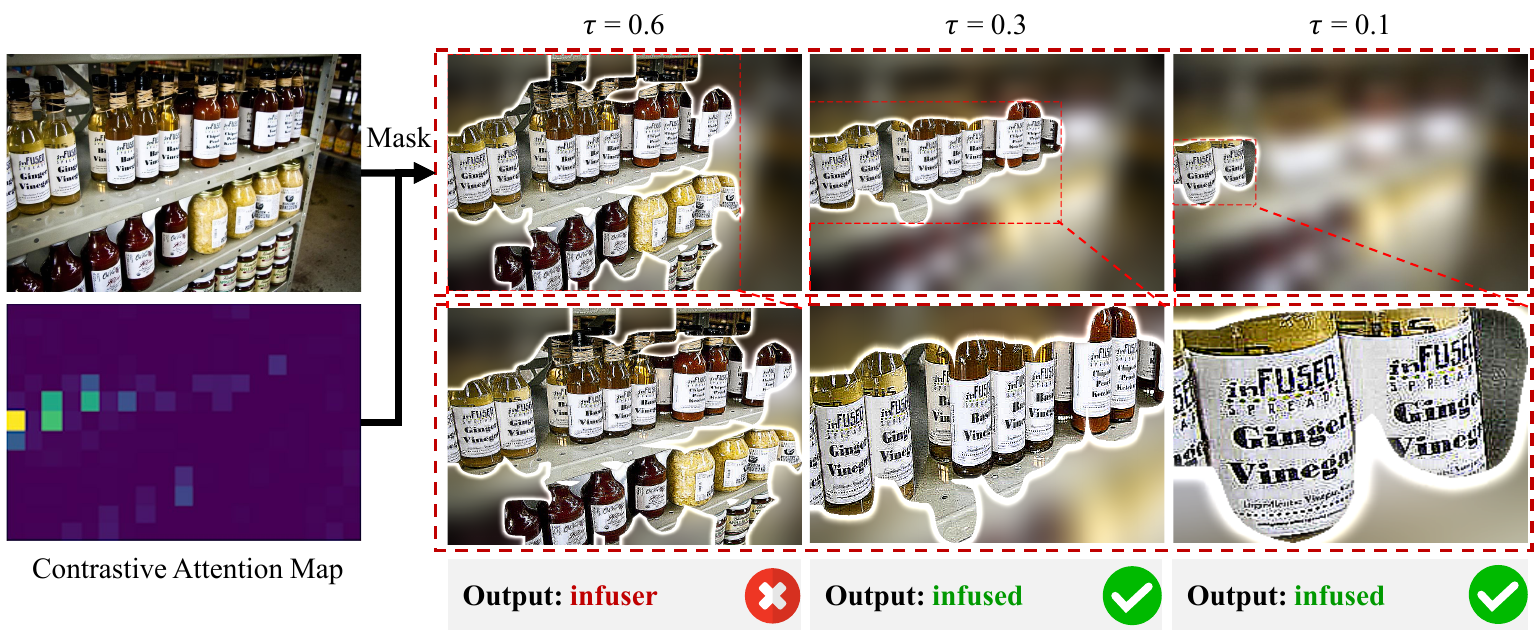}
    \caption{Visualization of contrastive attention refinement in VQA.
    As the threshold $\tau$ decreases from 0.6 to 0.1, CARVE progressively suppresses visual noise to isolate task-relevant regions.
    Precise grounding corrects initial hallucinations (Red $\times$) and yields accurate answers (Green $\checkmark$).}
    \label{fig:vqa_vis}
\end{figure}

\paragraph{Sensitivity Analysis of Mask Generation.}
Fig.~\ref{fig:heatmaps} reports accuracy on a 1,000-instance TextVQA subset across varying $p$ and $K$.
Setting $p \in [0.2, 0.6]$ with $K \in \{2, 3\}$ yields optimal performance by balancing noise suppression with object preservation.
Fig.~\ref{fig:vqa_vis} qualitatively confirms this: as $\tau$ tightens, CARVE progressively filters background noise, shifting focus to task-relevant cues and correcting hallucinations.

\begin{table}[t]
    \centering
    \caption{Component ablation (accuracy \%, \qwenSB). Direct $A_i^{(Q)}$ masking omits contrastive normalization; in-attention modification replaces pixel-level masking with direct attention weight manipulation.}
    \label{tab:component_ablation}
    \renewcommand{\arraystretch}{1}
    \resizebox{\textwidth}{!}{
    \setlength{\tabcolsep}{12pt}
    \begin{tabular}{lccc}
        \toprule
        \textbf{Method} & \textbf{TextVQA} & \textbf{ChartQA} & \textbf{ScienceQA} \\
        \midrule
        \cellcolor{gray!20}Vanilla & \cellcolor{gray!20}75.0 & \cellcolor{gray!20}79.12 & \cellcolor{gray!20}81.21 \\
        Direct $A_i^{(Q)}$ masking & \secondmetric{78.6}$_{\scriptscriptstyle\blacktriangle\,4.80}$ & \secondmetric{79.92}$_{\scriptscriptstyle\blacktriangle\,1.01}$ & \secondmetric{82.30}$_{\scriptscriptstyle\blacktriangle\,1.34}$ \\
        In-attention modification & 77.1$_{\scriptscriptstyle\blacktriangle\,2.80}$ & 79.56$_{\scriptscriptstyle\blacktriangle\,0.56}$ & 82.10$_{\scriptscriptstyle\blacktriangle\,1.10}$ \\
        \cellcolor{ours}\textbf{CARVE} (full) & \cellcolor{ours}\goodmetric{82.0}$_{{\color{ggreen}\scriptscriptstyle\blacktriangle}\,9.33}$ & \cellcolor{ours}\goodmetric{79.96}$_{{\color{ggreen}\scriptscriptstyle\blacktriangle}\,1.06}$ & \cellcolor{ours}\goodmetric{82.50}$_{{\color{ggreen}\scriptscriptstyle\blacktriangle}\,1.59}$ \\
        \bottomrule
    \end{tabular}
    }
\end{table}

\paragraph{Component Ablation Study.}
Table~\ref{tab:component_ablation} evaluates CARVE's core design choices using \qwenSB.
On TextVQA, direct $A_i^{(Q)}$ masking yields +4.80\% while CARVE achieves +9.33\%, confirming that contrastive normalization effectively suppresses $\mathcal{F}_{\text{vis}}(\mathcal{I})$.
In-attention modification produces the weakest gains, indicating that pixel-level masking outperforms internal representation manipulation.
Even on high-baseline tasks such as ChartQA and ScienceQA, all variants yield positive gains ranging from +0.56\% to +1.59\%, indicating that both contrastive normalization and pixel-level masking contribute.

\begin{table*}[t]
    \centering
    \caption{CARVE vs.\ external tools and ViCrop on TextVQA with \llavaSB: accuracy (\%) and inference latency per sample (s).}
    \label{tab:externals_time}
    \renewcommand{\arraystretch}{1}
    \resizebox{\textwidth}{!}{
        \scalebox{1}{
        \begin{tabular}{lcccccccc}
        \toprule
        \multirow{2}{*}{\textbf{Method}} & \multirow{2}{*}{\textbf{Original}} & \multirow{2}{*}{\textbf{SAM}} & \multirow{2}{*}{\textbf{YOLO}} & \multirow{2}{*}{\textbf{CLIP}} & \multicolumn{3}{c}{\textbf{ViCrop}} & \multirow{2}{*}{\textbf{CARVE}} \\
        \cmidrule(lr){6-8}
        & & & & & \texttt{rel-att} & \texttt{grad-att} & \texttt{pure-grad} & \\
        \midrule
        Accuracy & 47.80 & 49.42 & 48.84 & 48.55 & 55.17 & \secondmetric{56.06} & 51.67 & \goodmetric{58.2} \\
        \midrule
        Inf. Latency & 0.17 & 3.33 & 0.35 & 1.07 & 1.16 & 0.89 & 2.36 & 1.34 \\
        \bottomrule
        \end{tabular}
        }
    }
\end{table*}

\begin{table}[t]
    \centering
    \caption{GPU time (ms) and peak GPU memory (GB), averaged over 100 ChartQA samples. CARVE GPU time includes general description inference, attention processing, and enhanced inference.}
    \label{tab:profiling}
    \renewcommand{\arraystretch}{1}
    \resizebox{\textwidth}{!}{
    \setlength{\tabcolsep}{6pt}
    \begin{tabular}{l|cc|cc}
        \toprule
        \textbf{Model} & \textbf{Vanilla} & \textbf{CARVE} & \textbf{Vanilla Mem.} & \textbf{CARVE Mem.} \\
        \midrule
        \qwenTB  & 358 & 1,038 & 7.5  & 8.1 \\
        \qwenSB  & 415 & 1,286 & 15.9 & 16.6 \\
        \llavaSB   & 149 & 506 & 13.6 & 15.2 \\
        \llavaffB  & 233 & 762 & 25.7 & 28.9 \\
        \bottomrule
    \end{tabular}
    }
\end{table}

\paragraph{Comparison with External Tools.}
As shown in Table~\ref{tab:externals_time}, CARVE outperforms all external tool-based approaches: SAM~\cite{kirillov2023segment}, YOLO~\cite{redmon2016you}, CLIP~\cite{radford2021learning}, and the recent ViCrop~\cite{zhang2025mllms} variants.
External tools rely on generic segmentation algorithms that are not conditioned on the input question. While ViCrop effectively reduces visual noise through strategic cropping, it does not perform pixel-level noise masking. In terms of latency, CARVE requires an inference latency\footnote{Inference latency denotes end-to-end time from input to output, including all processing steps such as image preprocessing, GPU computation, and post-processing.} of 1.34 seconds, exceeding simpler approaches such as YOLO at 0.35 seconds but remaining within practical deployment constraints.

\paragraph{Inference Overhead Analysis.}
Table~\ref{tab:profiling} reports GPU time\footnote{GPU time measures pure GPU computation time, excluding image pre/post-processing and data transfer.} and memory measurements.
CARVE's GPU time ranges from 0.5 to 1.3 seconds per sample. Besides the three autoregressive passes, overhead arises from attention extraction and storage at each decoding step, connected component analysis for mask generation, and image cropping with resizing.
The memory overhead is primarily attributed to attention map storage during the first two passes, ranging from +0.6~GB to +3.2~GB and scaling with model size.
Early termination and attention caching strategies can further reduce this overhead.

\paragraph{Limitations.}
CARVE suppresses background to magnify spatially localized cues, making it most effective for local fine-grained perception such as text recognition and small-target identification. For tasks that hinge on \emph{global} spatial context---such as depth ordering or object-relationship reasoning---masking may remove necessary contextual cues; in such cases CARVE gracefully degrades to the vanilla model as $p \to 1.0$. CARVE also requires three forward passes, trading additional inference cost for accuracy.

\section{Conclusion}
\label{sec:conclusion}
In this work, we demonstrate that visual complexity correlates with attention entropy, which in turn negatively impacts VLMs' performance. Theoretically, under a multiplicative decomposition assumption, we show that contrasting attention maps between general and specific instructions approximately separates visual signal into semantic and noise components. To this end, we propose \textbf{CARVE}, a training-free method that exploits this decomposition to extract task-relevant signal through attention contrasting and pixel-level masking, yielding consistent gains across seven benchmarks.

\bibliography{reference}
\bibliographystyle{splncs04}

\newpage
\appendix

\section{Definition and Explanation}
\label{app:notation}

\subsection{Definition}

Table~\ref{tab:notation} summarizes the notation used throughout the paper.

\begin{table}[ht]
\centering
\caption{Notation used throughout the paper.}
\label{tab:notation}
\renewcommand{\arraystretch}{1.05}
    \setlength{\tabcolsep}{12pt}
    \resizebox{\textwidth}{!}{%
\begin{tabular}{l l l}
\toprule
    \textbf{Symbol} & \textbf{Definition} & \textbf{Description} \\
\midrule
    $\mathcal{I}\in\mathbb{R}^{H\times W\times 3}$ & Input image & Image with height $H$ and width $W$ \\
    $H, W$ & Image dimensions & Height and width in pixels \\
    $Q$ & Task-specific question & Task-specific question \\
    $G$ & General instruction & General instruction \\
    $A_{l,t}^{(Q)}\in\mathbb{R}^{N_v}$ & Question attention map & Attention map at layer $l$, step $t$ \\
    $A_{l,t}^{(G)}\in\mathbb{R}^{N_v}$ & General attention map & General question attention map \\
    $N_v$ & Visual tokens & Number of visual tokens \\
    $\mathcal{L}$ & Layer range & Layer indices \\
    $\mathcal{T}$ & Time range & Generation time step \\
    $t_{\text{end}}$ & Final step & Final generation step \\
    $\mathcal{H}(\cdot)$ & Shannon entropy & Attention distribution entropy \\
    $\overline{\mathcal{H}}$ & Overall entropy & Layer-averaged attention entropy \\
    $\mathcal{T}_c(\mathcal{I})$ & Texture complexity & Edge density from Canny detection \\
    $\mathcal{C}_c(\mathcal{I})$ & Color complexity & Hue diversity measure \\
    $\mathcal{E}(\mathcal{I})\!\in\!\{0,1\}^{H\times W}$ & Edge map & Binary edge map from Canny \\
    $\Psi_{RGB\!\rightarrow\!HSV}$ & Color transform & RGB to HSV transformation operator \\
    $\zeta_{ij}$ & Hue value & Hue value at pixel $(i,j)$ \\
    $\rho_b$ & Hue proportion & Fraction of pixels in hue bin $b$ \\
    $B$ & Hue bins & Number of hue bins \\
    $\mathcal{F}_{\text{vis}}(\mathcal{I})\in\mathbb{R}_+^{N_v}$ & Visual noise factor & Image-inherent visual noise component \\
    $\mathcal{F}_{\text{sem}}(Q,\mathcal{I})\in\mathbb{R}_+^{N_v}$ & Semantic signal factor & Task-related semantic signal component \\
    $\mathbf{1}_{N_v}$ & Uniform vector & Vector of ones \\
    $\hat{A}\in\mathbb{R}_+^{N_v}$ & Estimated attention & Estimated semantic attention \\
    $\lambda>0$ & Regularization & Regularization parameter \\
    $p\in(0,1]$ & Top-p percentile & Percentile threshold for masking \\
    $K\in\mathbb{N}$ & Max regions & Maximum regions to preserve \\
    $w_{t}$ & Temporal weights & Later token weighting with $w_t = t - t_{\text{start}} + 1$ \\
    $S\in\mathbb{R}^{H\times W}$ & Fused map & Spatially reshaped attention map \\
    $\mathcal{Q}_p(\cdot)$ & Percentile function & Top-$p$ percentile operator \\
    $\tau$ & Threshold & Computed threshold value \\
$M^*\subseteq\{1..H\}\times\{1..W\}$ & Final mask & Union of top-$K$ regions \\
    $R_k$ & Connected region & Connected component from thresholding \\
    $\Phi(\mathcal{I},M)$ & Visual extraction & Masking, cropping and resizing \\
    $\pi_{H\times W}$ & Spatial reshape & Token to image projection \\
    $\Xi$ & Attention extractor & Function to extract attention maps \\
    $\mathcal{M}$ & VLM & Vision-language model \\
    $L_{\text{total}}$ & Total layers & Number of model layers \\
    $N_q$ & Text tokens & Number of query tokens \\
\bottomrule
    \end{tabular}%
    }
\end{table}

\subsection{Explanation}
\label{subsec:explanation}

\begin{itemize}
\item \textbf{Time Step ($t$):} In the autoregressive generation process of vision-language models, a time step denotes the sequential position index in the output token sequence. The model generates responses token-by-token, where $t=1$ corresponds to the first generated token and $t=t_{\text{end}}$ represents the final token. At each time step, the model produces an attention distribution $A_{l,t}^{(Q)} \in \mathbb{R}^{N_v}$ over visual tokens.

\item \textbf{Visual Complexity ($\mathcal{T}_c(\mathcal{I})$, $\mathcal{C}_c(\mathcal{I})$):} Visual complexity quantifies the inherent characteristics of an image that can interfere with VLMs' attention mechanisms, decomposed into two orthogonal dimensions. Texture complexity $\mathcal{T}_c(\mathcal{I}) \in [0,1]$ measures the density of edge information using Canny edge detection, where higher values indicate more intricate patterns, object boundaries, and structural details. Color complexity $\mathcal{C}_c(\mathcal{I}) \in [0,1]$ captures the diversity of hue distribution in HSV color space through Shannon entropy, where higher values reflect greater chromatic variation.

\item \textbf{Visual Tokens ($N_v$):} Visual tokens constitute the discrete representational units obtained after processing an input image through a visual encoder. An image of dimensions $H \times W$ is partitioned and encoded into $N_v$ visual tokens, which form the fundamental units for visual information processing. The attention mechanism allocates weights across these $N_v$ tokens to determine which image regions to attend to.

\item \textbf{Semantic Signal Factor ($\mathcal{F}_{\text{sem}}(Q, \mathcal{I})$):} The semantic signal factor represents the question-specific component in the attention decomposition framework, valued in $\mathbb{R}^{N_v}_+$. This factor quantifies the semantic signal between each visual token and the given question $Q$. Under general instructions $G$ (\eg, ``describe this image''), this factor approximates a uniform distribution ($\mathcal{F}_{\text{sem}}(G, \mathcal{I}) \approx \mathbf{1}_{N_v}$), whereas task-specific questions yield elevated values in semantically relevant regions.

\item \textbf{Visual Noise Factor ($\mathcal{F}_{\text{vis}}(\mathcal{I})$):} The visual noise factor captures the image-inherent, question-independent attention component, valued in $\mathbb{R}^{N_v}_+$. This factor, determined by texture complexity and color diversity of the image, reflects the influence of visual content characteristics on attention distribution. Under general instructions, the attention distribution is predominantly governed by this factor: $A_{l,t}^{(G)}(\mathcal{I}) \approx \mathcal{F}_{\text{vis}}(\mathcal{I})$.
\end{itemize}

\section{Implementation Details}
\label{sec:implementation_details}

We conduct our experiments on servers equipped with NVIDIA RTX A6000 GPUs (for the main experiments in Tables 1--3) and NVIDIA H20 GPUs (for the extended evaluation in Tables 4--6). $\lambda$ is set to 0.05.
In practical implementation, CARVE requires three inference passes; however, the first two passes (extracting general instruction and task-specific question) can be terminated early. Specifically, when we require attention maps only from layers $\mathcal{L} = [L_{start}, L_{end}]$, the first two inference processes can halt upon completing layer $L_{end}$ computation, eliminating the need for full $L_{\text{total}}$ layer forward propagation. The third inference must run completely to generate the final answer.

\subsection{Extended Dataset Configuration}

For the extended evaluation (DocVQA, ChartQA, ScienceQA), we use the following configurations:
DocVQA uses 1,000 randomly sampled instances from the validation split (seed=42).
ChartQA uses the full test split (2,500 instances).
ScienceQA uses test-split instances that contain images (2,017 instances).
For the component ablation (Table 4 in the main paper), TextVQA uses 500 randomly sampled instances; ChartQA and ScienceQA use full test sets.
All extended experiments employ a unified hyperparameter configuration: $\mathcal{L} = [20, 25]$, $\mathcal{T} = [0, 5)$ with position weighting, $p = 0.3$, and $K = 1$.

\subsection{Evaluation Metrics}

\paragraph{Exact Match (EM).} A prediction is scored 1 if it exactly matches any ground-truth answer, and 0 otherwise. We report EM for DocVQA and ChartQA.

\paragraph{ANLS (Average Normalized Levenshtein Similarity).} For a prediction $\hat{a}$ and ground-truth set $\{a_1, \ldots, a_M\}$, the sample-level score is:
\begin{equation}
\text{ANLS}(\hat{a}, \{a_j\}) = \max_{j=1}^{M} s(\hat{a}, a_j), \quad
s(\hat{a}, a_j) = \begin{cases} 1 - \text{NL}(\hat{a}, a_j) & \text{if } \text{NL}(\hat{a}, a_j) < 0.5 \\ 0 & \text{otherwise} \end{cases}
\end{equation}
where $\text{NL}(\hat{a}, a_j) = d_{\text{Lev}}(\hat{a}, a_j) / \max(|\hat{a}|, |a_j|)$ is the normalized Levenshtein distance. The dataset-level ANLS averages over all samples. We report ANLS for DocVQA.

\paragraph{Relaxed Accuracy.} A prediction $\hat{a}$ is considered correct if it exactly matches the ground truth, or if both can be parsed as numbers and $|v(\hat{a}) - v(a)| \leq 0.05 \cdot |v(a)|$:
\begin{equation}
\text{RelaxedAcc}(\hat{a}, a) = \begin{cases} 1 & \text{if } \hat{a} = a \\ 1 & \text{if } |v(\hat{a}) - v(a)| \leq 0.05 \cdot |v(a)|, \; v(\cdot) \in \mathbb{R} \\ 0 & \text{otherwise} \end{cases}
\end{equation}
This $\pm$5\% tolerance accounts for rounding differences in numerical chart answers. We report Relaxed Accuracy for ChartQA.

\paragraph{Multiple-Choice Accuracy (MC Acc).} For ScienceQA, we use option-letter matching: the model's generated text is parsed for the selected option letter, and accuracy is computed over the test set.

\subsection{Robustness to Decoding Strategy}

All main experiments employ greedy decoding, following standard practice in visual reasoning evaluation. To investigate CARVE's sensitivity to decoding strategy, we evaluate \qwenTB with CARVE on a 200-instance TextVQA subset under varying sampling temperatures.

\begin{table}[ht]
    \centering
    \setlength{\tabcolsep}{12pt}
    \begin{tabular}{lccc}
        \toprule
        \textbf{Temperature} & \textbf{0.3} & \textbf{0.6} & \textbf{1.0} \\
        \midrule
        Accuracy (\%) & 78.5 & 77.0 & 77.0 \\
        \bottomrule
    \end{tabular}
    \caption{CARVE accuracy under different sampling temperatures (\qwenTB, 200-instance TextVQA subset).}
    \label{tab:temperature}
\end{table}

As shown in Table~\ref{tab:temperature}, performance remains stable across temperatures, with only a 1.5\% difference between near-greedy ($T$=0.3) and fully stochastic ($T$=1.0) decoding. This indicates that CARVE's improvements are robust to the choice of decoding strategy and are not an artifact of greedy decoding.

\section{Proofs and Additional Theorems}
\label{sec:theoretical_foundations}

The results in this section characterize the idealized model of Assumption~1; they are intended to motivate CARVE's design rather than to describe exact properties of real attention.

\subsection{Mathematical Basis of Attention Decomposition}

\textbf{Theorem C.1 (Existence and Conditional Uniqueness of Attention Decomposition):}
For any attention distribution $A_{l,t}^{(Q)}(\mathcal{I}) \in \mathbb{R}^{N_v}_+$, under the assumption $\mathcal{F}_{\text{sem}}(G, \mathcal{I}) = \mathbf{1}_{N_v}$, there exists a unique decomposition:
\begin{equation}
A_{l,t}^{(Q)}(\mathcal{I}) = \mathcal{F}_{\text{vis}}(\mathcal{I}) \otimes \mathcal{F}_{\text{sem}}(Q, \mathcal{I})
\end{equation}

\textit{Proof:} Define a logarithmic space mapping $\phi: \mathbb{R}_+ \rightarrow \mathbb{R}$ where $\phi(x) = \log(x)$. Under this transformation, the decomposition becomes additive in logarithmic space:
\begin{equation}
\phi(A_{l,t}^{(Q)}(\mathcal{I})) = \phi(\mathcal{F}_{\text{vis}}(\mathcal{I})) + \phi(\mathcal{F}_{\text{sem}}(Q, \mathcal{I}))
\end{equation}

Given the boundary condition that $\mathcal{F}_{\text{sem}}(G, \mathcal{I}) = \mathbf{1}_{N_v}$ when $Q = G$ (general instruction), we obtain:
\begin{equation}
\phi(\mathcal{F}_{\text{vis}}(\mathcal{I})) = \phi(A_{l,t}^{(G)}(\mathcal{I}))
\end{equation}

Consequently, through substitution:
\begin{equation}
\phi(\mathcal{F}_{\text{sem}}(Q, \mathcal{I})) = \phi(A_{l,t}^{(Q)}(\mathcal{I})) - \phi(A_{l,t}^{(G)}(\mathcal{I}))
\end{equation}

The unique solution is obtained via the inverse mapping $\phi^{-1}(x) = \exp(x)$. $\square$

\subsection{Convexity Analysis of the Optimization Problem}

\textbf{Theorem C.2 (Strict Convexity of Objective Function):}
The optimization objective
\begin{equation}
\mathcal{J}(\tilde{A}) = \sum_{i=1}^{N_v} \left( \tilde{A}_i \cdot A^{(G)}_i - A^{(Q)}_i \right)^2 + \lambda \sum_{i=1}^{N_v} \tilde{A}_i^2 \cdot A^{(G)}_i
\end{equation}
is strictly convex with respect to $\tilde{A}$.

\textit{Proof:} Computing the Hessian matrix reveals its structure. Since the objective function is separable across components $\tilde{A}_i$, the Hessian is diagonal with elements:
\begin{equation}
H_{ii} = \frac{\partial^2 \mathcal{J}}{\partial \tilde{A}_i^2} = 2(A^{(G)}_i)^2 + 2\lambda A^{(G)}_i = 2A^{(G)}_i(A^{(G)}_i + \lambda)
\end{equation}

Given that $A^{(G)}_i > 0$ and $\lambda > 0$, all diagonal elements are positive, thus $H \succ 0$ (positive definite). According to convex optimization theory, a twice continuously differentiable function with positive definite Hessian everywhere is strictly convex. $\square$

\subsection{Derivation and Uniqueness of Closed-form Solution}

\textbf{Theorem C.3 (Closed-form Expression of Optimal Solution):}
The optimization problem admits a unique global optimum:
\begin{equation}
\hat{A}_i = \frac{A^{(Q)}_i}{A^{(G)}_i + \lambda}
\end{equation}

\textit{Proof:} Applying first-order optimality conditions (KKT conditions):
\begin{equation}
\nabla_{\tilde{A}_i} \mathcal{J} = 2(\tilde{A}_i \cdot A^{(G)}_i - A^{(Q)}_i) \cdot A^{(G)}_i + 2\lambda \tilde{A}_i \cdot A^{(G)}_i = 0
\end{equation}

Rearranging terms yields:
\begin{equation}
\tilde{A}_i \cdot A^{(G)}_i \cdot (A^{(G)}_i + \lambda) = A^{(Q)}_i \cdot A^{(G)}_i
\end{equation}

Solving for $\tilde{A}_i$:
\begin{equation}
\tilde{A}_i = \frac{A^{(Q)}_i}{A^{(G)}_i + \lambda}
\end{equation}

By Theorem C.2's strict convexity, this solution represents the unique global optimum. $\square$

\subsection{Computational Optimization Potential of CARVE}

This section analyzes the computational optimization potential of the CARVE algorithm. While CARVE requires three inference passes, its structural properties enable significant optimization opportunities.

The key observation is that the first two inference passes (general instruction and task-specific question) only require extracting attention maps from intermediate layers, without completing full forward propagation or generating complete responses. This characteristic enables early termination strategies. Furthermore, the general attention maps $A^{(G)}$ depend solely on the input image and are independent of specific questions, creating opportunities for caching and reuse.

Let the forward propagation $\mathcal{P}: \mathbb{R}^{N_v} \rightarrow \mathbb{R}^{N_v}$ at layer $l$ have computational cost $c_l = \Theta(N_v^2)$. The baseline complexity without optimization is:
$$\mathcal{C}_{\text{baseline}} = 3L_{\text{total}} \cdot \Theta(N_v^2) + \Theta(|\mathcal{L}| \cdot |\mathcal{T}| \cdot N_v)$$

\paragraph{Early Termination Strategy.}

Since only attention maps from layers $\mathcal{L} = [L_{\text{start}}, L_{\text{end}}]$ are required, the first two inference passes can terminate after layer $L_{\text{end}}$:
$$\mathcal{C}_{\text{early}} = (2L_{\text{end}} + L_{\text{total}}) \cdot \Theta(N_v^2) + \Theta(|\mathcal{L}| \cdot |\mathcal{T}| \cdot N_v)$$

The relative computational savings rate is:
$$\eta_1 = \frac{\mathcal{C}_{\text{baseline}} - \mathcal{C}_{\text{early}}}{\mathcal{C}_{\text{baseline}}} = \frac{2(L_{\text{total}} - L_{\text{end}})}{3L_{\text{total}}} = \frac{2(1-\alpha)}{3}$$

where $\alpha = L_{\text{end}}/L_{\text{total}}$. For practical configurations with $\mathcal{L} = [20, 25]$ and $L_{\text{total}} = 28$, we have $\alpha = 25/28 \approx 0.89$, yielding theoretical savings of $\eta_1 \approx 7.1\%$.

\paragraph{Attention Caching Mechanism.}

The general attention maps $A^{(G)}$ depend only on the image $\mathcal{I}$ and can be reused across multiple questions. Define a cache mapping $\mathcal{H}: \mathcal{I} \rightarrow \{A^{(G)}_l\}_{l \in \mathcal{L}}$.

For $n$ different questions $\{Q_1, ..., Q_n\}$ on the same image, the total computational cost is:
$$\mathcal{C}_{\text{cached}}(n) = L_{\text{end}} \cdot \Theta(N_v^2) + n \cdot (L_{\text{end}} + L_{\text{total}}) \cdot \Theta(N_v^2)$$

compared to $3n \cdot L_{\text{total}} \cdot \Theta(N_v^2)$ for the baseline approach. The average cost per question becomes:
$$\bar{\mathcal{C}}_{\text{cached}} = \frac{L_{\text{end}}}{n} \cdot \Theta(N_v^2) + (L_{\text{end}} + L_{\text{total}}) \cdot \Theta(N_v^2)$$

As $n \to \infty$, the average cost approaches $(L_{\text{end}} + L_{\text{total}}) \cdot \Theta(N_v^2)$, yielding a speedup ratio relative to baseline:
$$S_{\text{cache}} = \frac{3L_{\text{total}}}{L_{\text{end}} + L_{\text{total}}} = \frac{3}{1 + \alpha}$$

For $\alpha = 0.89$, this gives $S_{\text{cache}} \approx 1.59$, representing approximately 37\% computational savings.

\paragraph{Combined Optimization Analysis.}

When processing batches containing repeated images, combining both strategies yields:
$$\mathcal{C}_{\text{combined}} = (1-\rho)L_{\text{end}} \cdot \Theta(N_v^2) + (L_{\text{end}} + L_{\text{total}}) \cdot \Theta(N_v^2)$$

where $\rho \in [0,1]$ denotes the cache hit rate. The relative speedup becomes:
$$S_{\text{combined}} = \frac{3L_{\text{total}}}{(2-\rho)L_{\text{end}} + L_{\text{total}}} = \frac{3}{(2-\rho)\alpha + 1}$$

Under practical scenarios with $\alpha = 0.89$ and $\rho = 0.3$, we obtain $S_{\text{combined}} \approx 1.19$, corresponding to approximately 16\% computational savings.

The space complexity remains $\mathcal{S}(\text{CARVE}) = \Theta(|\mathcal{L}| \cdot |\mathcal{T}| \cdot N_v)$. For typical configurations ($|\mathcal{L}| = 5$, $|\mathcal{T}| = 10$, $N_v = 1024$), this requires approximately 200KB of additional memory, which is negligible on modern hardware.

\section{Prompt Design}
\label{sec:prompt_design}

\begin{table}[ht]
    \centering
    \resizebox{\textwidth}{!}{%
    \begin{tabular}{lccc}
    \toprule
    \textbf{General Instruction} & \textbf{Accuracy (\%)} & \textbf{Std Dev (\%)} & \textbf{Relative Gain (\%)} \\
    \midrule
    Vanilla & 72.4 & 0.8 & -- \\
    ``Write a general description of the image.'' & \textbf{77.2} & \textbf{0.6} & \textbf{+6.63} \\
    ``Describe this image in detail.'' & 75.8 & 0.9 & +4.70 \\
    ``Provide a comprehensive overview of the image.'' & 75.2 & 1.4 & +3.87 \\
    ``What do you see in this image?'' & 74.9 & 1.2 & +3.45 \\
    ``Explain what appears in the image.'' & 74.8 & 1.7 & +3.31 \\
    \bottomrule
    \end{tabular}%
    }
    \caption{Comparison across general instructions.}
    \label{table:prompt}
\end{table}

To identify the optimal general instruction for inducing uniform attention distributions, we conducted experiments on a randomly sampled subset of 1000 instances from the TextVQA dataset using the \qwenTB.
Our objective was to identify prompts that encourage global image scanning without focusing on specific semantic regions.
To assess stability, we performed ten independent trials and computed standard deviations across runs.
To avoid discrepancies arising from layer and time step variations, we conduct experiments using $\mathcal{T}_{\text{full}}$ and $\mathcal{L}=[20, 25]$ as hyperparameters.
As shown in Table~\ref{table:prompt}, ``Write a general description of the image'' achieves both the highest accuracy (77.2\%) and the lowest standard deviation (0.6\%), indicating superior stability.
Meanwhile, ``What do you see in this image?'' and ``Explain what appears in the image.'' are excluded due to their poor stability.
Beyond considering accuracy and stability, we also need to consider the number of tokens generated by the VLM. Specifically, ``Describe this image in detail.'' and ``Provide a comprehensive overview of the image.'' are excluded because they output significantly more tokens than ``Write a general description of the image.''.
Based on the above considerations, we ultimately adopt ``Write a general description of the image.'' as the general instruction for CARVE.

\section{Datasets}
\label{datasets}
For A-OKVQA~\cite{schwenk2022aokvqa}, we utilize the validation split containing 1,145 questions across 1,122 images that require integrating visual perception with commonsense reasoning, evaluated using VQA-score accuracy. For POPE~\cite{li2023evaluating}, we employ 500 distinct images paired with 9,000 binary questions systematically designed to detect hallucination phenomena through polling-based object probing. For V$^*$~\cite{wu2024v}, we evaluate on 191 image-question pairs that demand fine-grained visual reasoning capabilities. For TextVQA~\cite{singh2019vqa}, we test on 3,166 images with 5,000 questions focusing on text comprehension abilities.

For TextVQA evaluation, we adopt the protocol established by Zhang~\etal~\cite{zhang2025mllms}, deliberately excluding OCR-extracted tokens from model inputs. We treat TextVQA identically to other visual reasoning benchmarks, providing only the image and question without auxiliary text annotations. While this configuration yields marginally reduced accuracy compared to OCR-augmented baselines in original implementations, it enables unbiased assessment of models' intrinsic visual text recognition capabilities, eliminating confounding factors from external OCR systems. This evaluation strategy ensures that performance metrics genuinely reflect the visual perception and text understanding abilities inherent to the vision-language models.

\section{Visualizations}
\label{sec:visualizations}

\begin{figure}[ht]
    \centering
    \begin{subfigure}[b]{0.18\textwidth}
        \includegraphics[width=\textwidth,height=3cm,keepaspectratio]{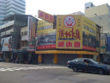}
        \caption{$\tau = 1.0$ }
    \end{subfigure}
    \hfill
    \begin{subfigure}[b]{0.18\textwidth}
        \includegraphics[width=\textwidth,height=3cm,keepaspectratio]{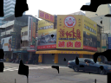}
        \caption{$\tau = 0.9$}
    \end{subfigure}
    \hfill
    \begin{subfigure}[b]{0.18\textwidth}
        \includegraphics[width=\textwidth,height=3cm,keepaspectratio]{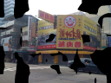}
        \caption{$\tau = 0.8$}
    \end{subfigure}
    \hfill
    \begin{subfigure}[b]{0.18\textwidth}
        \includegraphics[width=\textwidth,height=3cm,keepaspectratio]{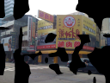}
        \caption{$\tau = 0.7$}
    \end{subfigure}
    \hfill
    \begin{subfigure}[b]{0.18\textwidth}
        \includegraphics[width=\textwidth,height=3cm,keepaspectratio]{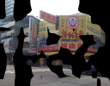}
        \caption{$\tau = 0.6$}
    \end{subfigure}

    \begin{subfigure}[b]{0.18\textwidth}
        \includegraphics[width=\textwidth,height=3cm,keepaspectratio]{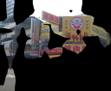}
        \caption{$\tau = 0.5$}
    \end{subfigure}
    \hfill
    \begin{subfigure}[b]{0.21\textwidth}
        \includegraphics[width=\textwidth,height=3cm,keepaspectratio]{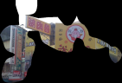}
        \caption{$\tau = 0.4$}
    \end{subfigure}
    \hfill
    \begin{subfigure}[b]{0.21\textwidth}
        \includegraphics[width=\textwidth,height=3cm,keepaspectratio]{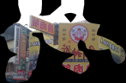}
        \caption{$\tau = 0.3$}
    \end{subfigure}
    \hfill
    \begin{subfigure}[b]{0.18\textwidth}
        \includegraphics[width=\textwidth,height=3cm,keepaspectratio]{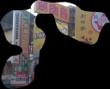}
        \caption{$\tau = 0.2$}
    \end{subfigure}
    \hfill
    \begin{subfigure}[b]{0.18\textwidth}
        \includegraphics[width=\textwidth,height=3cm,keepaspectratio]{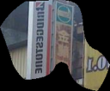}
        \caption{$\tau = 0.1$}
    \end{subfigure}

    \caption{Images masked with CARVE. The caption of each subfigure shows the threshold value $\tau$.}
    \label{fig:examples_2697_visualization}
\end{figure}

\begin{figure}[ht]
    \centering
    \newlength{\rowhgt}\setlength{\rowhgt}{0.135\textwidth}%
    \begin{subfigure}[b]{0.18\textwidth}
        \includegraphics[width=\textwidth,height=3cm,keepaspectratio]{figure/examples_121/crop_masked_image_1000_percent_1.000.jpg}
        \caption{$\tau = 1.0$}
    \end{subfigure}
    \hfill
    \begin{subfigure}[b]{0.18\textwidth}
        \includegraphics[width=\textwidth,height=3cm,keepaspectratio]{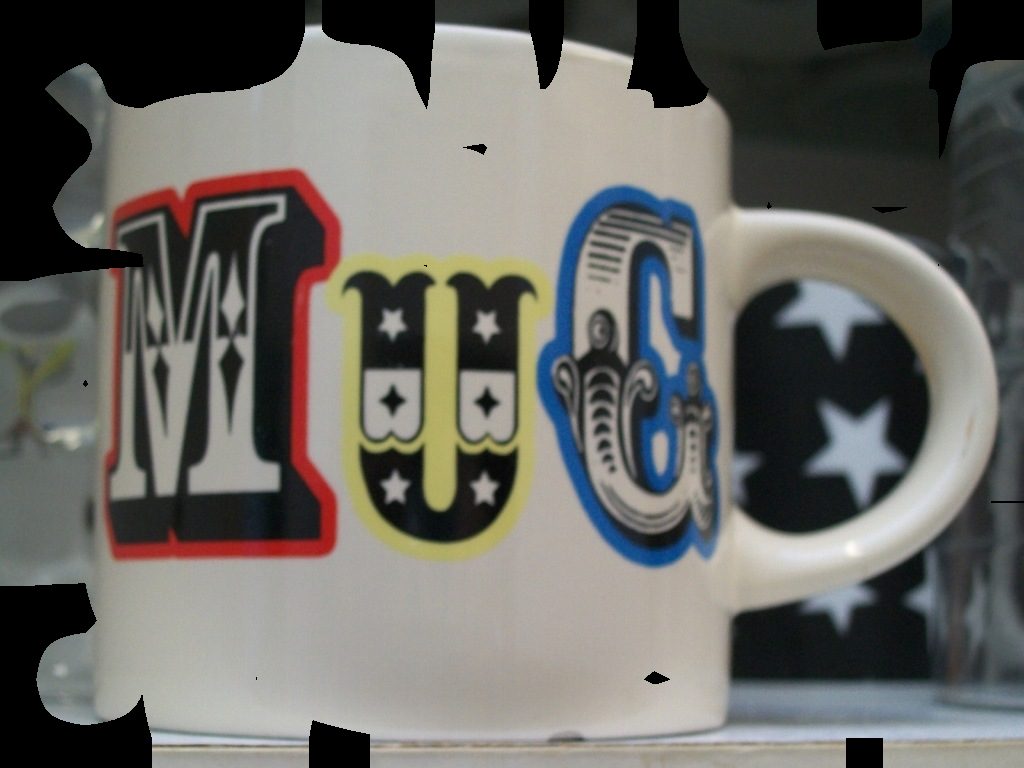}
        \caption{$\tau = 0.9$}
    \end{subfigure}
    \hfill
    \begin{subfigure}[b]{0.18\textwidth}
        \includegraphics[width=\textwidth,height=3cm,keepaspectratio]{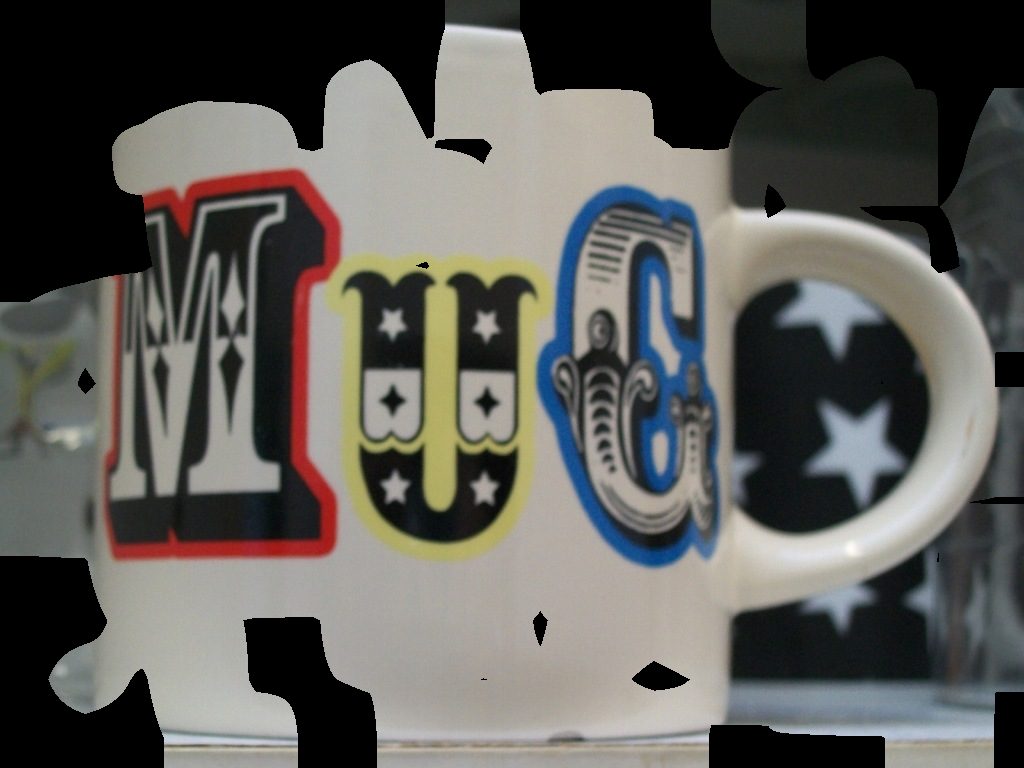}
        \caption{$\tau = 0.8$}
    \end{subfigure}
    \hfill
    \begin{subfigure}[b]{0.18\textwidth}
        \includegraphics[width=\textwidth,height=3cm,keepaspectratio]{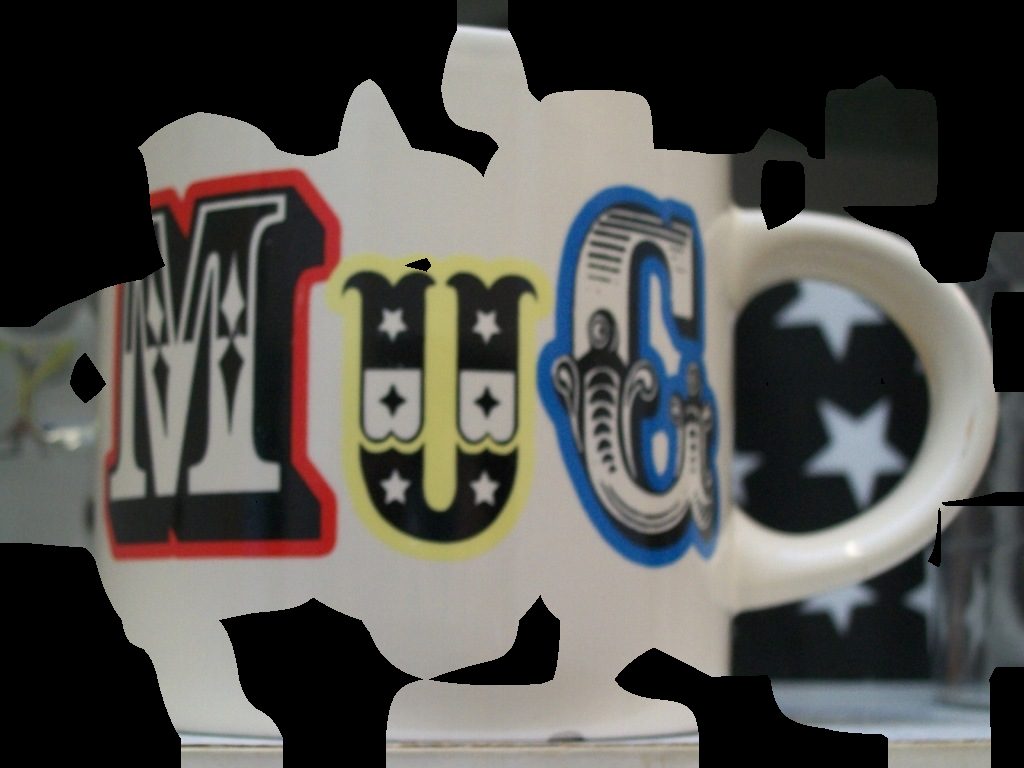}
        \caption{$\tau = 0.7$}
    \end{subfigure}
    \hfill
    \begin{subfigure}[b]{0.18\textwidth}
        \includegraphics[width=\textwidth,height=3cm,keepaspectratio]{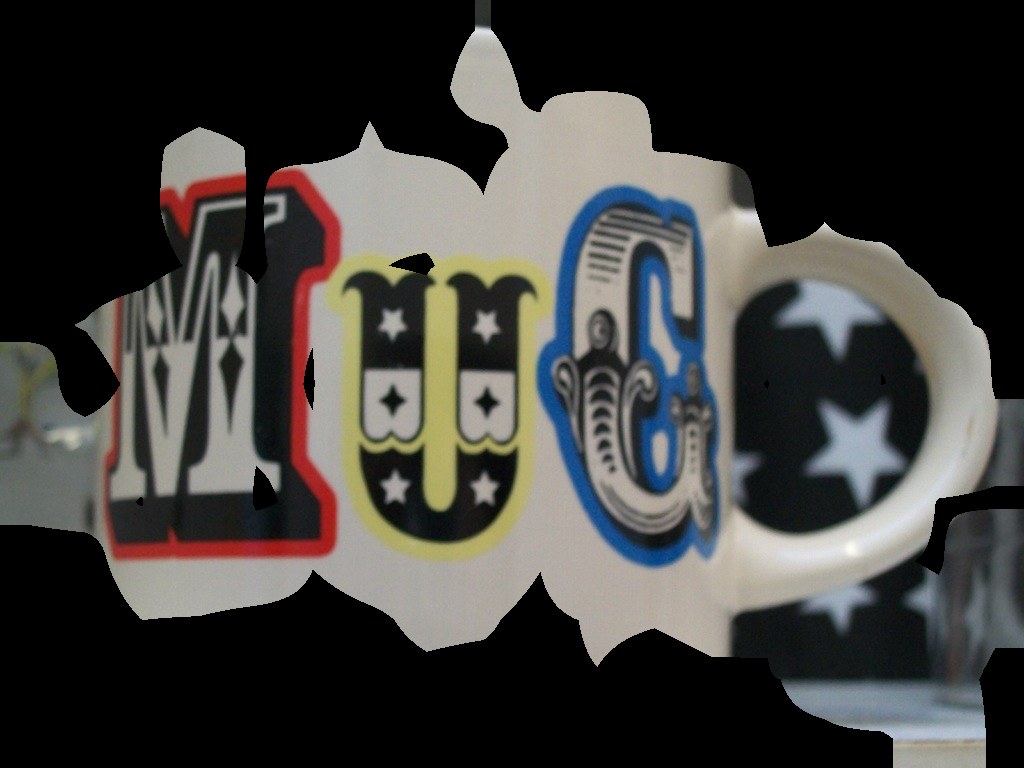}
        \caption{$\tau = 0.6$}
    \end{subfigure}

    \begin{subfigure}[b]{0.204\textwidth}
        \includegraphics[width=\textwidth,height=3cm,keepaspectratio]{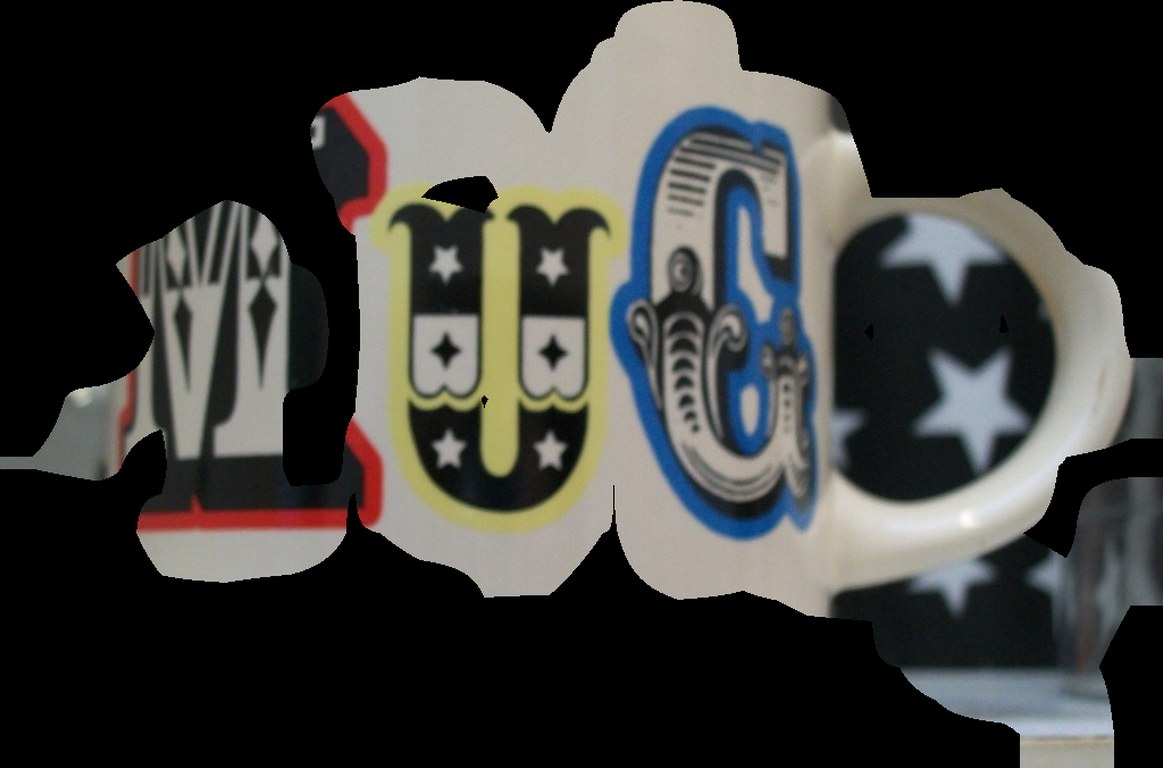}
        \caption{$\tau = 0.5$}
    \end{subfigure}
    \hfill
    \begin{subfigure}[b]{0.191\textwidth}
        \includegraphics[width=\textwidth,height=3cm,keepaspectratio]{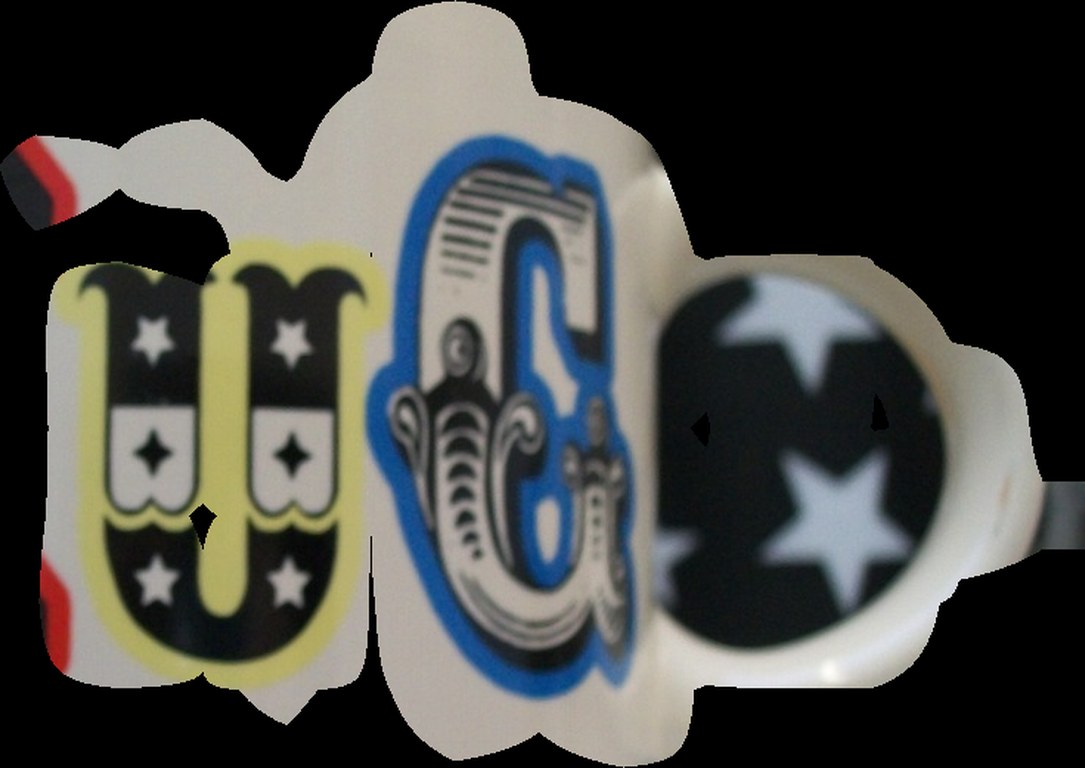}
        \caption{$\tau = 0.4$}
    \end{subfigure}
    \hfill
    \begin{subfigure}[b]{0.183\textwidth}
        \includegraphics[width=\textwidth,height=3cm,keepaspectratio]{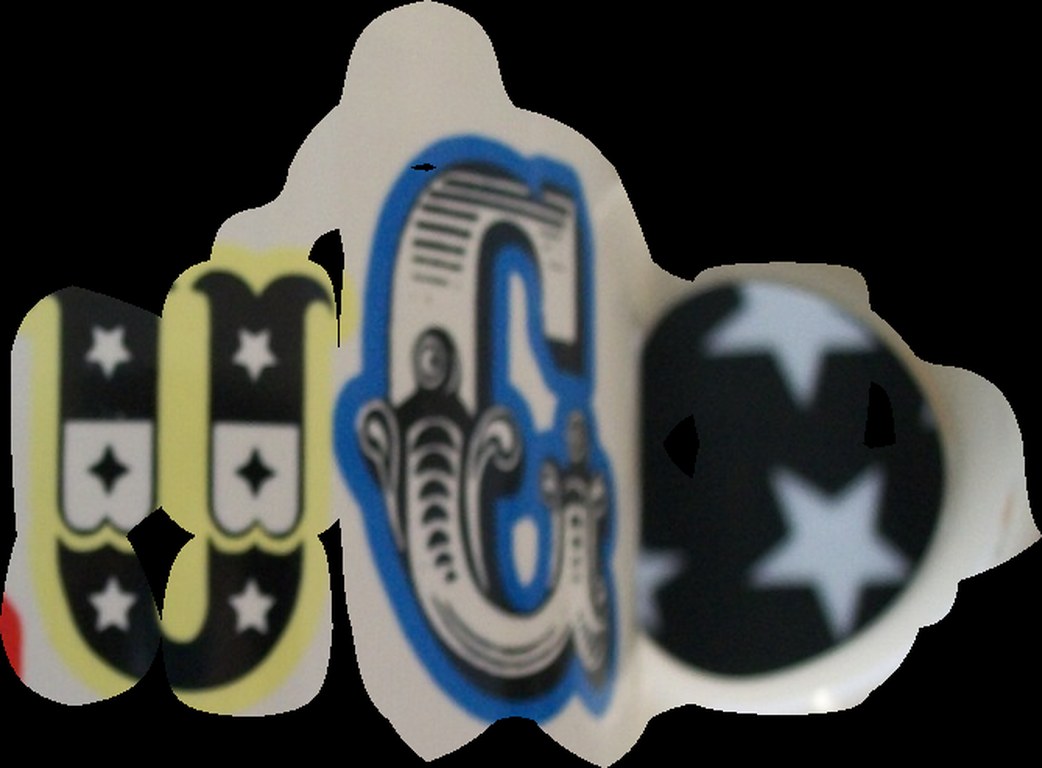}
        \caption{$\tau = 0.3$}
    \end{subfigure}
    \hfill
    \begin{subfigure}[b]{0.190\textwidth}
        \includegraphics[width=\textwidth,height=3cm,keepaspectratio]{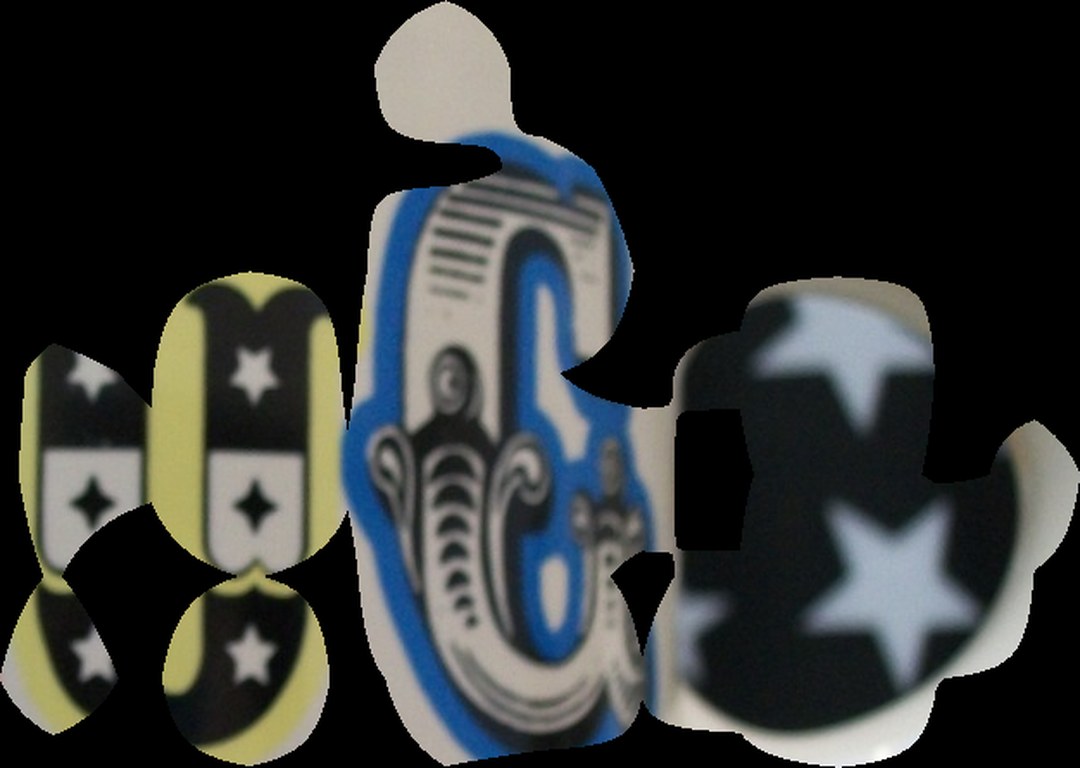}
        \caption{$\tau = 0.2$}
    \end{subfigure}
    \hfill
    \begin{subfigure}[b]{0.15\textwidth}
        \centering
        \includegraphics[height=\rowhgt]{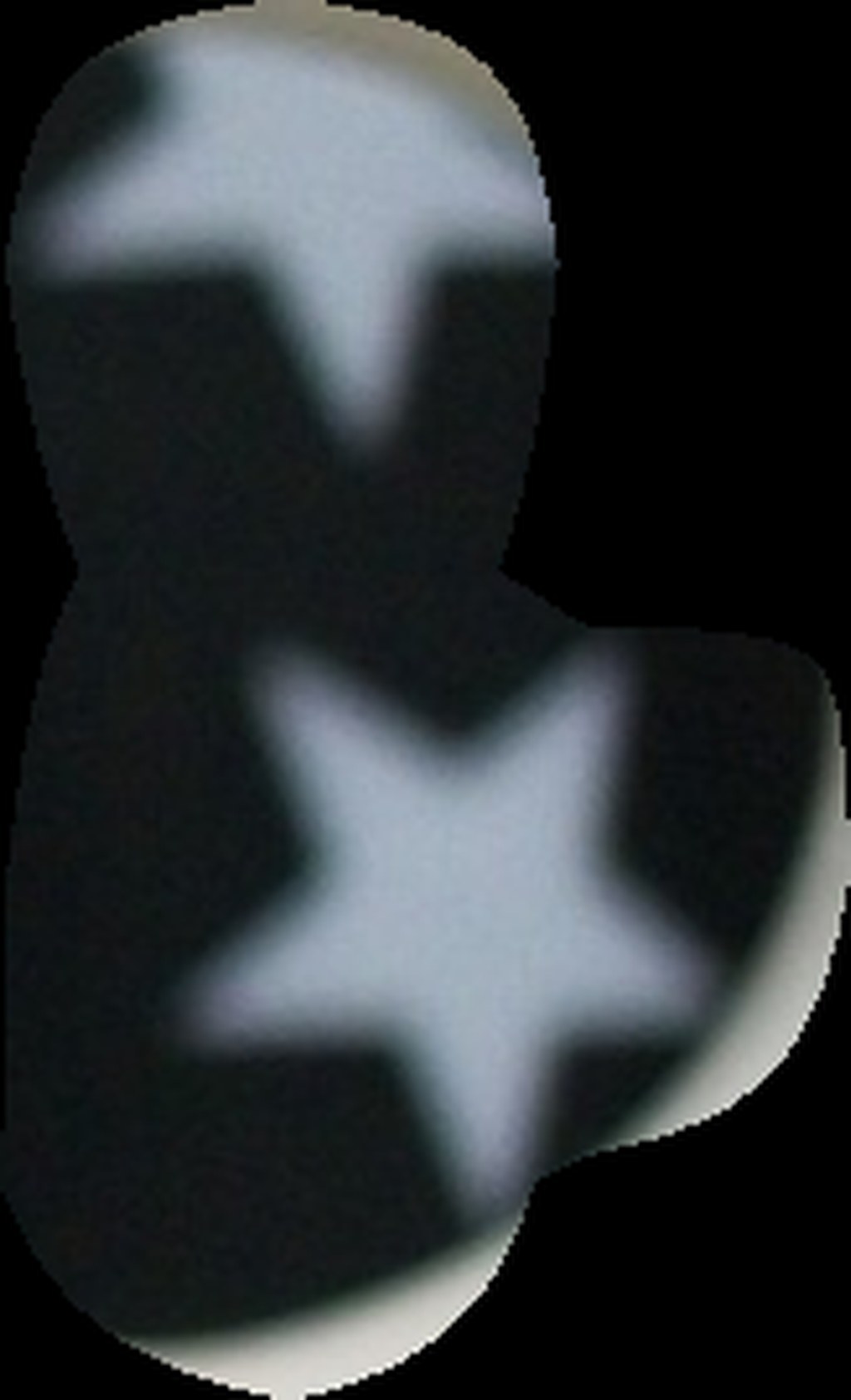}
        \caption{$\tau = 0.1$}
    \end{subfigure}

    \caption{Images masked with CARVE. The caption of each subfigure shows the threshold value $\tau$.}
    \label{fig:examples_121_visualization}
\end{figure}

This section presents visual analysis of masked images generated by CARVE across different threshold values $\tau$ from 1.0 (no masking) to 0.1 (aggressive masking). Figs.~\ref{fig:examples_2697_visualization} and~\ref{fig:examples_121_visualization} show two representative TextVQA samples where visual complexity initially causes incorrect predictions.

Fig.~\ref{fig:examples_2697_visualization} shows a street scene where the model fails to detect the Bridgestone sign at $\tau = 1.0$. Progressive masking removes background buildings and vehicles, enabling correct recognition at $\tau = 0.3$. In Fig.~\ref{fig:examples_121_visualization}, multiple decorative mugs cause shape misidentification through the cup handle. At $\tau = 0.2$, only the relevant mug remains, yielding the correct ``star'' answer. Across both samples, optimal performance occurs within $\tau \in [0.2, 0.4]$, where contrastive attention effectively preserves semantic signal while eliminating visual distractors.

\section{Decomposition Validation}
\label{app:decomposition_validation}

This section provides empirical validation for two modeling choices underlying CARVE: the multiplicative form of the attention decomposition and the uniformity assumption on general-instruction attention.

\subsection{Multiplicative vs.\ Additive Decomposition}

We compare CARVE's multiplicative contrast $\hat{A}_i = A_i^{(Q)} / (A_i^{(G)} + \lambda)$ against the additive alternative $\hat{A}_i = A_i^{(Q)} - A_i^{(G)}$. Both variants share identical downstream pipeline (fusion, thresholding, masking). Experiments are conducted with \qwenSB using the unified configuration ($\mathcal{L} = [20,25]$, $\mathcal{T}_{\text{full}}$, $p = 0.3$, $K = 1$).

\begin{table}[ht]
    \centering
    \caption{Multiplicative vs.\ additive decomposition (accuracy \%, \qwenSB).}
    \label{tab:mult_vs_add}
    \setlength{\tabcolsep}{10pt}
    \begin{tabular}{lccc}
        \toprule
        \textbf{Decomposition} & \textbf{TextVQA} & \textbf{ChartQA} & \textbf{ScienceQA} \\
        \midrule
        \cellcolor{gray!20}Vanilla & \cellcolor{gray!20}75.0 & \cellcolor{gray!20}79.12 & \cellcolor{gray!20}81.21 \\
        Additive ($A^{(Q)} - A^{(G)}$) & 78.4 & 79.84 & 81.90 \\
        \textbf{Multiplicative (CARVE)} & \textbf{82.0} & \textbf{79.96} & \textbf{82.50} \\
        \bottomrule
    \end{tabular}
\end{table}

As shown in Table~\ref{tab:mult_vs_add}, additive contrast yields consistent but weaker improvements across all benchmarks. The performance gap is most pronounced on TextVQA, where heterogeneous visual noise magnitudes across regions render subtraction ineffective: high-noise regions retain large residuals that dominate the fused attention map. Multiplicative normalization eliminates this scale dependence by treating visual noise as a per-token gain factor, consistent with Assumption~1.

\subsection{Uniformity of General-Instruction Attention}

To validate the assumption $\mathcal{F}_{\text{sem}}(G, \mathcal{I}) \approx \mathbf{1}_{N_v}$, we measure the KL divergence of attention distributions from the uniform distribution $\mathcal{U} = (1/N_v, \ldots, 1/N_v)$ over 500 randomly sampled TextVQA images using \qwenSB at layers $\mathcal{L} = [20, 25]$.

\begin{table}[ht]
    \centering
    \caption{KL divergence from uniform distribution ($\mathcal{U}$). Lower values indicate closer approximation to uniformity. Statistics are computed over 500 images.}
    \label{tab:kl_uniformity}
    \setlength{\tabcolsep}{10pt}
    \begin{tabular}{lcc}
        \toprule
        \textbf{Attention Source} & \textbf{Mean KL}$(\cdot \| \mathcal{U})$ & \textbf{Std} \\
        \midrule
        General instruction $A^{(G)}$ & 0.147 & 0.094 \\
        Task-specific question $A^{(Q)}$ & 0.812 & 0.271 \\
        \bottomrule
    \end{tabular}
\end{table}

Table~\ref{tab:kl_uniformity} confirms that general-instruction attention distributions lie substantially closer to uniformity than their task-specific counterparts ($0.147 \ll 0.812$), with more than a five-fold difference. This supports the modeling assumption that general instructions induce near-uniform semantic factors, enabling their attention maps to serve as proxies for the visual noise component $\mathcal{F}_{\text{vis}}(\mathcal{I})$.

\section{Statistical Analysis of V$^*$ Results}
\label{app:vstar_statistical}

V$^*$~\cite{wu2024v} contains 191 image--question pairs specifically designed to evaluate fine-grained visual search in cluttered scenes. Given the modest sample size, we provide statistical validation of the observed improvements.

\subsection{Bootstrap Confidence Intervals}

We estimate 95\% confidence intervals via non-parametric bootstrap with $B = 10{,}000$ resamples. For each resample, $n = 191$ predictions are drawn with replacement and accuracy is computed. Table~\ref{tab:vstar_bootstrap} reports bootstrap percentile intervals.

\begin{table}[ht]
    \centering
    \caption{Bootstrap 95\% confidence intervals on V$^*$ accuracy (\%). $B = 10{,}000$ resamples.}
    \label{tab:vstar_bootstrap}
    \setlength{\tabcolsep}{6pt}
    \begin{tabular}{lccc}
        \toprule
        \textbf{Model} & \textbf{Method} & \textbf{Accuracy} & \textbf{95\% CI} \\
        \midrule
        \multirow{2}{*}{\qwenTB}
        & Vanilla & 50.3 & [42.9, 57.6] \\
        & CARVE & 57.1 & [49.7, 64.4] \\
        \midrule
        \multirow{2}{*}{\qwenSB}
        & Vanilla & 50.8 & [43.5, 58.1] \\
        & CARVE & 59.7 & [52.4, 66.0] \\
        \midrule
        \multirow{2}{*}{\llavaSB}
        & Vanilla & 38.7 & [31.9, 46.1] \\
        & CARVE & 66.5 & [59.7, 72.8] \\
        \midrule
        \multirow{2}{*}{\llavaffB}
        & Vanilla & 42.4 & [35.6, 49.7] \\
        & CARVE & 74.3 & [67.5, 80.1] \\
        \bottomrule
    \end{tabular}
\end{table}

\subsection{McNemar's Test}

To further assess paired significance, we apply McNemar's test on the $2 \times 2$ contingency table of per-sample correctness changes, as reported in Table~\ref{tab:vstar_mcnemar}.

\begin{table}[ht]
    \centering
    \caption{McNemar's test on V$^*$. $b$: vanilla-correct $\to$ CARVE-incorrect; $c$: vanilla-incorrect $\to$ CARVE-correct.}
    \label{tab:vstar_mcnemar}
    \setlength{\tabcolsep}{8pt}
    \begin{tabular}{lccc}
        \toprule
        \textbf{Model} & $b$ & $c$ & $p$-value \\
        \midrule
        \qwenTB  & 11 & 24 & $< 0.05$ \\
        \qwenSB  & 13 & 30 & $< 0.05$ \\
        \llavaSB   & 18 & 71 & $< 0.001$ \\
        \llavaffB  & 14 & 75 & $< 0.001$ \\
        \bottomrule
    \end{tabular}
\end{table}

All models yield statistically significant improvements ($p < 0.05$), with LLaVA models achieving $p < 0.001$. The large V$^*$ gains are expected given the benchmark's design: every sample demands precise localization of a small target among abundant visual distractors---the exact scenario where contrastive attention-based noise suppression provides the greatest benefit. The Qwen models, equipped with stronger visual encoders, exhibit smaller but still significant gains, indicating partial inherent robustness to visual clutter.

\section{Evaluation on Recent VLMs}
\label{sec:recent_vlms}

The main paper evaluates CARVE on the Qwen2.5-VL and LLaVA-1.5 families. To verify that CARVE generalizes to more recent architectures, we apply the identical training-free pipeline---without any modification or per-model tuning---to three additional models: \textsc{Qwen3-VL-2B}, \textsc{Qwen3-VL-4B}, and \textsc{Gemma-3-27B-it}. We report accuracy on TextVQA under the same protocol used in the main paper (image and question only, no external OCR), with the unified configuration $\mathcal{L} = [20, 25]$, $\mathcal{T}_{\text{full}}$, $p = 0.3$, $K = 1$.

\begin{table}[ht]
    \centering
    \caption{CARVE on recent VLMs (TextVQA accuracy \%). The same training-free pipeline is applied without per-model tuning.}
    \label{tab:recent_vlms}
    \setlength{\tabcolsep}{10pt}
    \begin{tabular}{lccc}
        \toprule
        \textbf{Model} & \textbf{Vanilla} & \textbf{CARVE} & \textbf{$\Delta$ (\%)} \\
        \midrule
        \textsc{Qwen3-VL-2B} & 79.9 & \goodmetric{84.2} & +5.38 \\
        \textsc{Qwen3-VL-4B} & 81.8 & \goodmetric{86.7} & +5.99 \\
        \textsc{Gemma-3-27B-it} & 58.9 & \goodmetric{60.2} & +2.21 \\
        \bottomrule
    \end{tabular}
\end{table}

As shown in Table~\ref{tab:recent_vlms}, CARVE yields consistent positive gains on all three recent models. The Qwen3-VL family, despite stronger baselines than Qwen2.5-VL, still benefits substantially (+5.38\% and +5.99\%), indicating that the visual-noise interference identified in the main paper persists in newer architectures and that contrastive attention focusing remains effective. Gemma-3-27B-it, which uses a different visual encoder and a substantially larger language backbone, also improves (+2.21\%), suggesting that CARVE transfers across the model families and scales we evaluate.

\section{Limitations}
\label{sec:limitations}

\begin{figure}[ht]
    \centering
    \begin{subfigure}[b]{0.40\textwidth}
        \includegraphics[width=\textwidth]{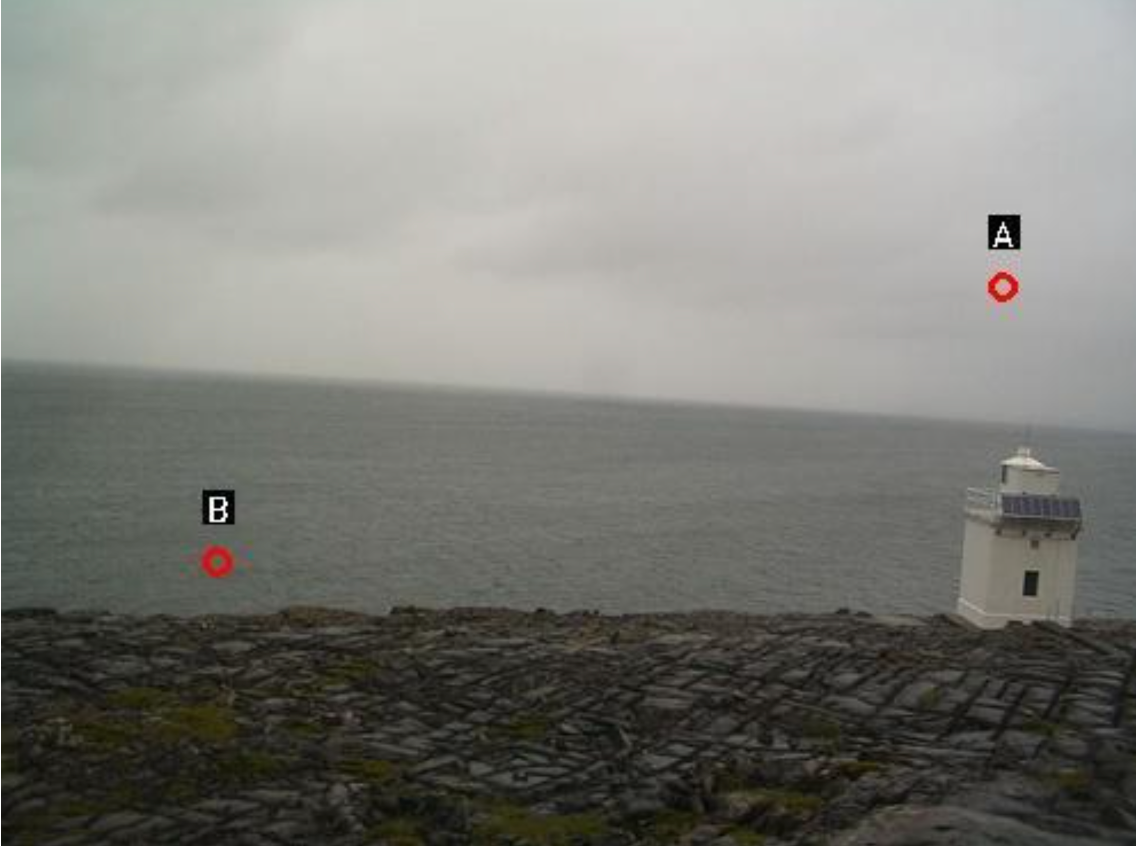}
        \caption{Original image}
        \label{fig:blink_original}
    \end{subfigure}
    \hfill
    \begin{subfigure}[b]{0.436\textwidth}
        \includegraphics[width=\textwidth]{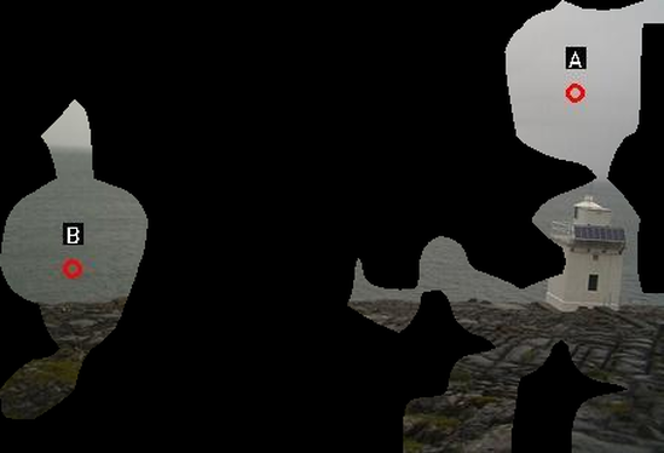}
        \caption{After CARVE masking}
        \label{fig:blink_carve}
    \end{subfigure}
    \caption{Failure case on a global spatial reasoning task (BLINK relative depth). CARVE precisely retains the regions around the probed points A and B, but removing the surrounding background eliminates the global depth cues (the continuous foreground-to-horizon ground plane) needed to determine which point is closer to the camera.}
    \label{fig:blink_limitation}
\end{figure}

\paragraph{Task Scope.}
CARVE enhances visual reasoning by masking background regions, cropping, and magnifying semantic regions, making it well-suited for tasks where task-relevant information is spatially localized, such as text recognition (TextVQA, DocVQA), chart reading (ChartQA), and fine-grained target identification (V$^*$).
For tasks that require global spatial reasoning---such as depth estimation or spatial relationship judgment---masking background regions may remove contextual cues necessary for correct inference.
Fig.~\ref{fig:blink_limitation} illustrates this boundary on a BLINK~\cite{fu2024blink} relative-depth sample (``Which point is closer to the camera, A or B?''). CARVE correctly localizes the two probed points, but the masking severs the continuous ground plane that connects the near foreground to the far horizon---precisely the global cue required for depth ordering---causing the depth judgment to fail.
When $p \to 1.0$, CARVE degrades to the vanilla model by retaining the full image, introducing no additional harm.
The current formulation addresses single-image, single-question scenarios; extending contrastive attention to multi-image reasoning~\cite{fu2026contextnav} and video understanding~\cite{fu2026videostir} remains an avenue for future work.

\paragraph{Decomposition Assumption.}
Assumption~1 models attention as a multiplicative product of visual noise and semantic signal: $A^{(Q)} = \mathcal{F}_{\text{vis}} \otimes \mathcal{F}_{\text{sem}}$.
In practice, the attention mechanism computes $\text{softmax}(QK^\top / \sqrt{d})$, where nonlinear interactions exist between query and visual features; the multiplicative independence assumption is a simplification of this process.
The uniformity approximation of general-instruction attention ($\text{KL}(A^{(G)} \| \mathcal{U}) = 0.147$) may weaken on images with high visual complexity.
The component ablation (Table~4 in the main paper) shows that contrastive normalization significantly outperforms direct masking (+9.33\% vs.\ +4.80\% on TextVQA), validating the effectiveness of this modeling choice in practice.

\paragraph{Vanilla Model Capability Dependency.}
CARVE's gains are positively correlated with the visual perception demands of the task.
On tasks where visual noise is prominent (e.g., TextVQA, V$^*$), CARVE yields robust and significant improvements; on tasks dominated by knowledge reasoning (e.g., ScienceQA), gains are smaller but remain positive, indicating that CARVE does not impair non-visual reasoning capabilities.
When a gap exists between the vanilla model's capacity and the target task (e.g., LLaVA-1.5 on document understanding), CARVE still provides consistent improvements, though the magnitude is bounded by the model's inherent capability ceiling.

\paragraph{Inference Overhead.}
As reported in Table~6 of the main paper, CARVE requires three inference passes, with GPU time approximately three times that of the vanilla model.
Beyond the inference passes themselves, attention map extraction and storage, image thresholding, and connected component analysis of the mask contribute additional overhead.
Given CARVE's broad applicability as a training-free method and its consistent improvements on visual reasoning performance, this overhead is acceptable.
The early termination and attention caching strategies analyzed in Sect.~\ref{sec:theoretical_foundations} can further reduce computational cost.
Distilling CARVE's masking decisions into a lightweight prediction head to enable single-pass inference is a promising future direction.

\end{document}